\pdfoutput=1

\documentclass[11pt]{article}

\usepackage[final]{acl}

\usepackage{times}
\usepackage{latexsym}

\usepackage[T1]{fontenc}

\usepackage[utf8]{inputenc}

\usepackage{microtype}

\usepackage{comment}
\usepackage{adjustbox}
\usepackage{amsmath}
\usepackage{inconsolata}
\usepackage{multirow}
\usepackage{booktabs}
\usepackage{subcaption}
\usepackage{graphicx}
\usepackage{listings}
\usepackage{tcolorbox}

\newcounter{promptno}[section]
\newlength\mystoreparindent
\newenvironment{prompt}[1][]
{
  \setlength{\mystoreparindent}{\the\parindent}
  \setlength{\parindent}{0pt}
  \refstepcounter{promptno}
  \par\medskip
  \noindent
  \begin{tcolorbox}[left=1pt,right=1pt]
  \textsc{{template \small\thesubsection.\thepromptno}}\\
  \small
  \tt
}{
  \end{tcolorbox}
  \setlength{\parindent}{\mystoreparindent}
  \medskip
}

\usepackage{xcolor}
\usepackage{colortbl}
\usepackage{array}
\usepackage{fancyvrb}

\lstdefinelanguage{json}{
  basicstyle=\ttfamily\scriptsize,
  breaklines=true,
  breakatwhitespace=false,
  columns=flexible,  
  showstringspaces=false,
  numbers=none,
  frame=single,
  backgroundcolor=\color{gray!10},
}

\usepackage{tikz}

\newcommand{\clemtodd}{\texttt{clem:todd}}

\title{{\tt clem:todd}: A Framework for the Systematic Benchmarking of\\ LLM-Based
Task-Oriented Dialogue System Realisations

}

\author{%
Chalamalasetti Kranti${^\mathbf{1}}$, Sherzod Hakimov${^\mathbf{1}}$, David Schlangen${^\mathbf{1,2}}$\\$^{\mathbf{1}}$Computational Linguistics, Department of Linguistics\\
University of Potsdam, Germany\\
$^{\mathbf{2}}$German Research Center for Artificial Intelligence (DFKI), Berlin, Germany\\
{\texttt{\{kranti.chalamalasetti, sherzod.hakimov, david.schlangen\}@uni-potsdam.de}}
}

\begin{document}
\maketitle
\begin{abstract}
The emergence of instruction-tuned large language models (LLMs) has advanced the field of dialogue systems, enabling both realistic user simulations and robust multi-turn conversational agents. However, existing research often evaluates these components 
in isolation, either focusing on a single user simulator or a specific system design, limiting the generalisability of insights across architectures and configurations. In this work, we propose
\clemtodd\ 
(chat-optimized LLMs for task-oriented dialogue systems development), a flexible framework for systematically evaluating dialogue systems under consistent conditions. \clemtodd\ enables detailed benchmarking across combinations of user simulators and dialogue systems, whether existing models from literature or newly developed ones. 
To the best of our knowledge, \clemtodd\ is the first evaluation framework for task-oriented dialogue systems that supports plug-and-play integration and ensures uniform datasets, evaluation metrics, and computational constraints. We showcase \clemtodd’s flexibility by re-evaluating existing task-oriented dialogue systems within this unified setup and integrating three newly proposed dialogue systems into the same evaluation pipeline. Our results provide actionable insights into how architecture, scale, and prompting strategies affect dialogue performance, offering practical guidance for building efficient and effective conversational AI systems.
\end{abstract}

\section{Introduction}
\label{sec:introduction}
Task-oriented dialogue systems~\citep{McTear2002, DBLP:conf/ijcnlp/LiCLGC17, eric-etal-2017-key, DBLP:conf/emnlp/BudzianowskiWTC18,  DBLP:conf/acl/BalakrishnanRUW19, chen-etal-2019-working, elder-etal-2020-make} help users complete specific goals, such as booking travel or making reservations, through multi-turn natural language interactions. These systems must accurately interpret user intent, manage ambiguity, and respond coherently. 

\begin{figure}[ht]
  \vspace*{-.2cm}
  \includegraphics[width=0.48\textwidth]{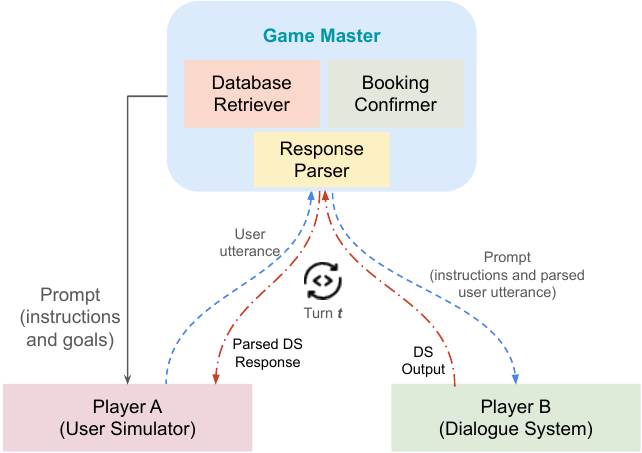}
  \caption{\clemtodd\ framework facilitates turn-based interactions between a user simulator (Player A) and a dialogue system (Player B), coordinated by a Game Master module.
  }
  \label{fig:ds_clemtod_details}
  \vspace*{-.7cm}
\end{figure}

With the rise of LLMs, 
these systems~\citep{DBLP:conf/sigdial/HudecekD23, DBLP:conf/acl/XuMYSH24} have significantly improved in 
handling task-oriented, goal-driven conversations. In addition to enhancing dialogue system capabilities, LLMs now play dual roles: 
as end-to-end dialogue systems~\citep{chung-etal-2023-instructtods, DBLP:conf/coling/DongCY25} and as user simulators~\citep{DBLP:conf/nips/KojimaGRMI22, DBLP:conf/slt/KaziLZHT24} for training and evaluation. However, evaluation remains a challenge~\citep{DBLP:conf/icmi/CasasTKMC20, DBLP:journals/corr/abs-2312-13871}. Current evaluations often suffer from inconsistent datasets, metrics, and compute settings, making it difficult to compare models or draw conclusions about system design.

A recent paradigm for evaluating LLMs as agents, involves self-play in conversational ``games''.
Frameworks such as clembench~\citep{DBLP:conf/emnlp/ChalamalasettiG23}, GameEval~\citep{DBLP:journals/corr/abs-2308-10032}, SPAG~\citep{DBLP:conf/nips/ChengHXZDHDL24}, and TextArena~\citep{guertler2025textarena} 
assess the capabilities of LLMs 
in instruction following, logical reasoning, and context retention across multi-turn settings. However, they do not address the evaluation of interactive systems built on LLMs, such as task-oriented dialogue (TOD) systems. 

To address this gap,
we introduce \clemtodd~\footnote{\url{https://github.com/clp-research/clem-todd}} (see Figure~\ref{fig:ds_clemtod_details}), an evaluation framework that builds on the self-play framework setup for TOD. The user simulator and dialogue system act as the two players in the self-play setup, with a central controller (game master) managing turn-taking, task flow, and interactions with external tools such as databases. This setup supports consistent benchmarking, stress testing, and detailed analysis under shared datasets, metrics, and resource constraints.

Using \clemtodd, we explore three 
research questions: (1) Can LLM-based self-play be adapted to evaluate interactive TOD systems? 
(2) On the MultiWOZ2.2 benchmark, a widely used dataset for multi-domain TOD, 
how do user simulators and dialogue systems perform across different models and architectures? 
(3) Can \clemtodd\ facilitate evaluation beyond MultiWOZ to plug in new instances, either derived from other datasets or generated synthetically, to enhance robustness testing?

We benchmark both existing dialogue systems and three newly developed ones within our framework,
providing systematic comparisons and insights across a range of configurations. Our contributions are as follows: (a) we propose \clemtodd, a self-play-based evaluation framework for system-level benchmarking of 
TOD systems; (b) we re-evaluate existing systems and our proposed dialogue systems under the same benchmarking setup; and (c) we analyse trade-offs across architectures, model sizes, and user simulators, with a focus on task performance and computational cost. 

\section{Related Work}
\label{sec:relatedwork}
Our work builds on research in: 
LLM-based 
TOD systems, 
user simulators, 
task evaluation, and self-play frameworks for multi-turn interaction.

\paragraph{LLM-Based TOD Systems} leverage LLMs in diverse architectural setups to enable successful task completion. 
\citet{DBLP:conf/sigdial/HudecekD23} proposed a modular system using LLMs across functional components, while InstructTOD~\citep{chung-etal-2023-instructtods} and ProTOD~\citep{DBLP:conf/coling/DongCY25} extended this design with performance-oriented enhancements. In contrast, AutoTOD~\citep{DBLP:conf/acl/XuMYSH24} introduced a zero-shot monolithic system, showcasing the potential of LLMs as end-to-end systems. DARD~\citep{DBLP:journals/corr/abs-2411-00427} proposed a multi-agent framework, coordinating domain-specific LLMs via a central LLM-based dialogue manager.

\paragraph{LLM-Based User Simulators} use LLMs to generate realistic, goal-driven user interactions for training and evaluating dialogue systems. \citet{DBLP:conf/nips/KojimaGRMI22, DBLP:journals/corr/abs-2309-13233} leveraged in-context learning to produce diverse and contextually appropriate responses. \citet{DBLP:journals/corr/abs-2402-13374} fine-tuned LLMs on domain-specific data to mitigate hallucinations, while \citet{DBLP:conf/slt/KaziLZHT24} proposed an adaptive simulator that dynamically responds to the dialogue system it interacts with. 

\paragraph{LLM-Based Task Evaluation} explores using LLMs as evaluators to assess dialogue quality and task success. 
Recent work has prompted LLMs to act as judges in both open-ended~\citep{DBLP:conf/nips/ZhengC00WZL0LXZ23} and task-specific contexts~\citep{DBLP:conf/emnlp/LiuIXWXZ23, DBLP:journals/corr/abs-2406-12624, DBLP:conf/emnlp/KimSLLSWNL0S24, DBLP:conf/iclr/ChanCSYXZF024,bavaresco2024llmsinsteadhumanjudges}. \citet{DBLP:conf/naacl/JiaKLPNSGK24} explored automatic dialogue evaluation using LLMs, and \citet{DBLP:conf/acl/HashemiERDK24} introduced LLM-Rubric, a framework for structured, rubric-based evaluation of language outputs.

\paragraph{Frameworks for Dialogue System Evaluation} \citet{10.5555/3454287.3455511}  propose an evaluation framework for dialogue-system self-play in open-domain dialogue systems. \citet{cheng-etal-2022-multiwoz} discuss the importance of using a goal-oriented user simulator for improved evaluation. \citet{DBLP:conf/acl/XuMYSH24} showcase the evaluation of different dialogue systems against a single user simulator. Additionally ConvLab~\citep{DBLP:conf/acl/ZhuZFLTLPGZH20} and ParlAI\footnote{\url{https://parl.ai/}} offer infrastructure for developing dialogue agents, while DialEvalMetrics~\citep{DBLP:journals/corr/abs-2106-03706} focuses on evaluation. However, none provide a unified setup for controlled, plug-and-play experimentation across LLM-based user simulators and dialogue systems, evaluated with consistent metrics and compute constraints. \clemtodd\ fills this gap by enabling systematic exploration of architectural choices, simulator–system interactions, and scaling trade-offs.

\paragraph{LLM-Based Self-Play Frameworks} Self-play has recently been used to evaluate LLMs on multi-turn tasks in frameworks like clembench~\citep{DBLP:conf/emnlp/ChalamalasettiG23}, GameEval~\citep{DBLP:journals/corr/abs-2308-10032}, SPAG~\citep{DBLP:conf/nips/ChengHXZDHDL24}, and TextArena~\citep{guertler2025textarena}. These systems focus on model-level evaluation, while \clemtodd\ adapts self-play for benchmarking full dialogue systems.
\\\\
\noindent
Building on these foundations, we propose \clemtodd\, an evaluation framework that integrates LLM-based user simulators, dialogue systems, and automatic evaluation. It supports flexible combinations of monolithic, modular, and multi-agent architectures and evaluates them under consistent metrics and compute settings. This design enables comprehensive benchmarking and reveals system interaction dynamics in controlled conditions.

\section {Methodology}
\label{sec:methodology}
We frame task-oriented dialogue as a form of ``Assistance Game''~\citep{laidlaw2024scalably}, where one player needs assistance completing a task, and the other must work toward recovering their reward function. While \citet{laidlaw2024scalably} address the learning problem, 
we use this framing to develop an evaluation framework for testing different approaches to building dialogue systems.

\subsection{``Self-Play'' of Task-Oriented Dialogue}
\label{sec:methodology-selfplay}
We build on \textit{clembench}~\citep{DBLP:conf/emnlp/ChalamalasettiG23}, a framework that 
realises conversational games
between two (LLM-simulated) players coordinated by a Game Master. \clemtodd\ reuses and modifies the two-player setup as: (a) Player A is an LLM-based user simulator that receives a goal and a prompt (see Appendix~\ref{fig:usimulator_prompt}) describing the task; (b) Player B is a dialogue system that receives user simulator utterances and a prompt (see Appendix~\ref{fig:monods_prompt} and~\ref{fig:modllm_ds_prompt}) 
describing the task, with the objective of satisfying Player A’s goal; 
and (c) the Game Master orchestrates the interaction by passing utterances between the two players and managing turn-level coordination.

The Game Master enforces stricter format constraints than the base \textit{clembench} setup by requiring adherence to a predefined Tool Schema (in JSON format) (\textit{e.g.}, \texttt{followup}, \texttt{querydb}, \texttt{validatebooking}) to reflect the demands of real-world deployment. Any violation aborts the conversation, reinforcing that instruction following and format compliance is essential. Valid tool calls are executed programmatically, either to retrieve database results, validate booking details, or forward follow-up messages to the user simulator. The schema is given in Figures~\ref{lst:tool-schema-1}-\ref{lst:tool-schema-6} in Appendix.

\begin{figure}[ht!]
  \includegraphics[width=0.48\textwidth]{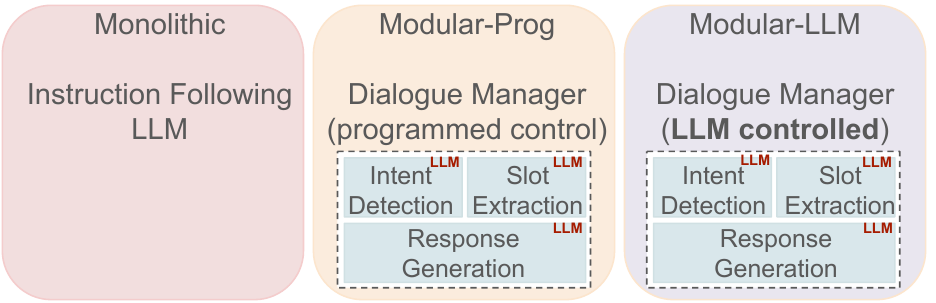}
  \caption{Overview of proposed dialogue system architectures. Monolithic, Modular-Prog, and Modular-LLM configurations vary in how control and components are handled by LLMs.}
  \label{fig:ds_proposed_systems}
\end{figure}

A full interaction (see Figure~\ref{fig:ds_clemtod_details}) proceeds as follows: the user simulator (Player A) is primed with a goal (\textit{e.g.}, booking a train ticket) and generates a natural language utterance (\textit{e.g.}, ``I am looking for a train to Paris from London''). This utterance is passed to the dialogue system (Player B) via the Game Master, which then returns a response (\textit{e.g.}, ``Do you have a specific day/time in mind?'') to the simulator. The loop continues until the conversation ends when Player A signals by either generating the \texttt{DONE} token or when the maximum turn limit ($15$) is reached. The Game Master then evaluates the task success by comparing the system’s outputs with the ground truth goal specification.

\clemtodd\ draws inspiration from ``self-play'' and adapts it to goal-driven task-oriented dialogue systems, where the user and system take on distinct roles. Building on prior works~\citep{DBLP:conf/acl/ZhuZFLTLPGZH20, DBLP:conf/acl/XuMYSH24, DBLP:conf/slt/KaziLZHT24}, \clemtodd\ employs a schema-constrained interaction loop and supports plug-and-play integration of user simulators and dialogue systems across different architectures, enabling more robust and extensible benchmarking.

\subsection{User Simulator}
We leverage the generative capabilities of LLMs to simulate realistic user interactions. The user simulator is prompted  (see Figure~\ref{fig:usimulator_prompt} in Appendix~\ref{sec:appendix-prompt-templates}) with a set of goals, represented as natural language objectives, and 
is intended to engage in a dialogue with the system under evaluation to fulfill them. The prompt includes relevant task context and dialogue history, enabling the user simulator to produce coherent, goal-oriented responses 
in the conversation. Such simulators facilitate automated evaluation workflows, complementing real users.

\subsection{Dialogue System Variants}
As shown in Figure~\ref{fig:ds_clemtod_details}, \clemtodd\ enables us to plug in various dialogue systems. We benchmark two existing systems and introduce three variants (Figure~\ref{fig:ds_proposed_systems}), each representing a different architectural paradigm and control mechanism.

\subsubsection{Existing Dialogue Systems}
We integrated two representative systems from prior work: AutoTOD, a zero-shot monolithic system that reported results on MultiWOZ 2.0 dataset~\citep{DBLP:conf/acl/XuMYSH24}, and 
a modular pipeline~\citep{DBLP:conf/sigdial/HudecekD23} that reported results on MultiWOZ 2.2 dataset. These systems were selected for their architectural diversity and publicly available implementations. Attempts to incorporate other approaches were not successful~\footnote{
InstructTOD, evaluted on Multiwoz 2.1~\citep{chung-etal-2023-instructtods}  
encountered compatibility issues between the agent implementation and LLM outputs, which led to repeated execution errors. For ProTOD, evaluted on Multiwoz 2.0~\citep{DBLP:conf/coling/DongCY25}, the publicly released code lacked essential implementation components, rendering integration into our framework infeasible.} (see Appendix~\ref{sec:appendix-existing-sys-integration} for details).

\subsubsection{Proposed Dialogue Systems}

\paragraph{Monolithic Dialogue System} This variant uses a single instruction-tuned LLM (see  Figure~\ref{fig:monods_prompt} for the prompt) to  
act as a the dialogue system, with backend actions performed via ``tool use''; i.e., prediction of API calls.
Although similar in style to the zero-shot approach proposed by~\citet{DBLP:conf/acl/XuMYSH24}, our implementation differs in that their method uses a LangChain-powered agent and generates raw SQL queries directly from the LLM, whereas ours avoids third-party agents and instead generates structured tool calls conforming to the Tool Schema as described in Section~\ref{sec:methodology-selfplay}.

\paragraph{Modular Dialogue System} This design decomposes the dialogue pipeline processing into sub-modules for intent detection, slot extraction, dialogue management, and response generation. (see Figures \ref{fig:modprog_intent_detection}, \ref{fig:modprog_slot_extraction}, \ref{fig:modprog_response_generation} in the Appendix for prompt templates).
We explore two variants that reflect distinct control strategies:  (a) \textit{Modular-Prog}, where a programmatic dialogue manager executes modules in a fixed order (a static pipeline); and (b) \textit{Modular-LLM}, where an LLM acts as the dialogue manager, (see Figure~\ref{fig:modllm_ds_prompt} for the prompt),
dynamically selecting the next sub-module based on intermediate outputs (an adaptive pipeline). All sub-modules 
use LLMs and produce 
outputs aligned with the Tool Schema. Outputs are validated before being passed downstream. 
Both variants follow a simple, traditional modular design, allowing us to assess how classical pipelines perform when augmented with LLMs.

\section {Experimental Setup}
\label{sec:expsetup}
\paragraph{Dataset} We conduct experiments using the widely used \textit{MultiWOZ 2.2} dataset~\citep{DBLP:conf/emnlp/BudzianowskiWTC18, DBLP:conf/naacl/HungLVPG22}, which contains annotated dialogues across multiple domains such as restaurant, hotel, and train booking. Importantly for our purposes, each dialogue comes with a verbally specified goal that was given to the `customer' (e.g., ``you want to book a Chinese restaurant in the south of the city, for a party of 4''), and a structured representation of the dialogue outcome. 
For evaluation, we use only the test split ($1000$ tasks).

To ensure a focus on end-to-end task completion, we filter the dataset to include only tasks that end in booking actions, resulting in $117$ tasks ($60$ single-domain and $57$ multi-domain dialogues) consisting of three domains (restaurant, hotel, and train). These filtered dialogues goals, after pre-processing to remove HTML artifacts, are directly used as input in the user simulator prompt. This filtering is specific to our evaluation goals and does not reflect a limitation of the \clemtodd\ framework, which supports arbitrary task types and domains (more details in Section~\ref{sec:clemtod-extendability}).

\paragraph{Model Selection} We experiment with both open-weight and closed-weight LLMs to compare their effectiveness in task-oriented dialogue settings. The open-weight models include \textit{Llama 3.1–8B}, \textit{Llama 3.2–1B}, \textit{Llama-3.2-3B}, \textit{Llama 3.3–70B} from the Llama family~\citep{llama31}, as well as Qwen2.5–7B, 32B~\citep{qwen25} 
allowing us to examine performance differences across model sizes and families. For closed-weight models, we evaluate GPT-4o (version \textit{gpt-4o-2024-08-06}). 
Additionally, we conducted preliminary tests with DeepSeek, but it was excluded from final experiments due to persistent issues with inconsistent JSON output formatting.

This model selection enables us to explore trade-offs between model accessibility, computational cost, and task success in dialogue system deployment. All models were run with a fixed \textit{temperature} of $0$ and a \textit{max\_new\_tokens} limit of $500$. GPT-4o was accessed via the OpenAI API,
while open-weight models were loaded and executed locally on A100 GPUs.

\paragraph{Evaluation Metrics}
Our evaluation considers both traditional~\citep{DBLP:conf/emnlp/BudzianowskiWTC18} and goal-oriented metrics for measuring the efficiency of the dialogue system. We use \textit{Inform}, which measures whether the system provides the correct entity matching the user's request; 
and \textit{Booking Accuracy}, a metric recommended by~\citep{DBLP:conf/acl/XuMYSH24}, which evaluates whether the final task such as making a reservation was successfully completed. These metrics are computed automatically by comparing system outputs against the ground-truth annotations in the MultiWOZ test set. Since we filtered tasks to include only those requiring booking actions in the restaurant, hotel, and train domains tasks, the resulting set excludes entity attributes such as phone number or address, which appear only in the attraction and taxi domains. Therefore, we do not report the \textit{Success} metric in our evaluation.

In addition to task-specific metrics, we also evaluate the overall quality of dialogue. Inspired by \citet{DBLP:conf/slt/KaziLZHT24}, we report dialogue-level metrics such as \textit{naturalness}, \textit{coherence}, and \textit{diversity}. Furthermore, to assess the realism of the user simulator’s utterances, we conduct a ``Turing test'' with human evaluators to determine their naturalness (see Appendix~\ref{sec:appendix-dlgeval-humaneval} for details).

\section{Results}
We organize our results around the research questions introduced in Section~\ref{sec:introduction}, providing quantitative comparisons and visualizations.

\subsection{Extending Self-Play to Task-Oriented Dialogue}
\textit{Can existing LLM-based self-play frameworks be extended to evaluate interactive systems built using LLMs, such as 
TOD systems?} We explore this question by building on the self-play framework, clembench~\citep{DBLP:conf/emnlp/ChalamalasettiG23}, 
to a 
TOD setting in a framework that we call \clemtodd. It retains the 
core structure of the existing framework: a two-agent interaction loop, a central controller for turn coordination, logging and scoring mechanisms for evaluation. In addition, we introduce: (i) interfacing with external APIs or simulated databases, and (ii) computing task-specific metrics such as \textit{slot coverage}, \textit{dialogue efficiency}, and \textit{task success}.

\begin{table}[t]
    \begin{minipage}[t]{0.48\textwidth}
      \vspace{0pt} 
      \scriptsize
        \begin{tabular}{lcccc}
          \toprule
          {\textbf{Model (US)}} & \textbf{Naturalness} & \textbf{Coherence} & \textbf{Diversity} & \textbf{TT}  \\
          \midrule
          \verb|Llama-3.2-1B|   & 3.84 & 2.84 & 1.00 & - \\
          \verb|Llama-3.2-3B|   & 2.42 & 1.75 & 1.00 & - \\
          \verb|Qwen2.5-7B|    & 2.75 & 2.42 & 1.05 & - \\
          \verb|Llama-3.1-8B|   & 2.82 & 2.18 & 1.02 & - \\
          \verb|Qwen2.5-32B|   & \textbf{4.65} & 2.95 & 1.00 & \textbf{0.38} \\
          \verb|Llama-3.3-70B|  & 4.42 & \textbf{2.97} & 1.00 & 0.16 \\
          \bottomrule          
        \end{tabular}
        \caption{Dialogue quality comparison of user simulators for the Monolithic architecture-based dialogue system (using the model: \texttt{Qwen2.5-32B}), evaluated on Naturalness (N), Coherence (C), and Dialogue Diversity (D) metrics using \texttt{GPT-4o}. The Turing Test (TT) metric measures the percentage of dialogues from a random sample of $50$ that are judged as human-like, based on human evaluation. Higher scores indicate better performance.}
        \label{tab:usdlgquality}
  \end{minipage}  
\end{table}

To validate 
its viability, we instantiated \clemtodd\ 
with different pairings of user simulators (US) and dialogue systems (DS) 
starting with an examination of 
US effectiveness and its impact on dialogue evaluation.
To address this overarching question, we first attend to the sub-questions.

\subsubsection{Analyzing the Role of User Simulators}
\label{subsec:usquality}
\textit{Which LLMs produce the most effective and realistic user simulation behavior in terms of dialogue naturalness and coherence?} 
This question is central to reliable evaluation~\citep{DBLP:journals/ker/PietquinH13, DBLP:journals/corr/abs-2309-13233, DBLP:journals/corr/abs-2402-13374} in 
TOD, where systems rely on multi-turn interactions to achieve user-defined goals.

To identify user simulators that show coherent and natural behavior, we evaluate dialogue systems of varying capacities ($32$B and $70$B) paired with simulators from the Qwen and LLaMA model families, ($1$B $\sim $ $70$B) on single-domain test dialogues.

We assess dialogue quality using both automatic and human evaluations. For automatic evaluation, we adopt the LLM-as-a-judge framework~\citep{DBLP:conf/slt/KaziLZHT24}, \verb|using GPT-4o| in a zero-shot setting to rate generated user utterances. 
Results in Table~\ref{tab:usdlgquality} indicate that both \textit{naturalness} and \textit{coherence} generally improve with increasing model size, while \textit{dialogue-level diversity} remains relatively constant (
$\sim 1.0$). These trends are corroborated by parallel evaluations using the open-weight \verb|Llama-3.3-70B| model as the judge (see Table~\ref{tab:usdlgquality_l370b}).

To further validate these findings, we conducted a ``Turing Test'' (see Appendix~\ref{sec:appendix-dlgeval-humaneval}) where an annotator 
chose the more natural dialogue 
between an LLM output and its corresponding ground-truth. This evaluation assesses the dialogues produced by \texttt{Qwen2.5-32B} and \texttt{LLaMA-3.3-70B} relative to ground-truth dialogues for the same task, focusing specifically on perceived naturalness.

We developed a simple user interface (see Figure~\ref{fig:usturingtest_interface} in Appendix) that displays two dialogues side by side, allowing annotators to select which dialogue appears more natural. In each comparison, one dialogue is generated by either \verb|Qwen2.5-32B| or \verb|LLaMA-3.3-70B|, and the other is the corresponding ground-truth dialogue from the corpus~\citep{DBLP:conf/emnlp/BudzianowskiWTC18}.

As shown in Table~\ref{tab:usdlgquality}, $19$ out of $50$ randomly sampled dialogues generated by \verb|Qwen2.5-32B| were preferred ($0.38$) over the ground truth,
compared to $8$ ($0.16$) from \verb|LLaMA-3.3-70B|. This 
result highlights \verb|Qwen2.5-32B|’s strength in generating realistic user behavior.

\subsubsection{Robustness of Dialogue Systems to User Simulator Variability}
\textit{How robust are dialogue systems when interacting with user simulators of varying capability and coherence?}
Although dialogue quality assessment (in Section~\ref{subsec:usquality}) provides insight into 
simulated dialogues, it is also important to evaluate the robustness of dialogue systems to user simulators for reliable evaluation. As shown in Table~\ref{tab:usvsdsperf}, the overall performance of dialogue systems generally is sensitive to the choice (and with this, as established by Table~\ref{tab:usdlgquality}, the quality) of the user simulator, but to differing extents: The dialogue systems 
realised with \texttt{Qwen2.5-32B} appear to be more capable of maintaining task success despite differences in simulator behavior. These systems can recover from incoherent user turns and use clarifying strategies 
to manage imperfect inputs and maintain goal alignment. In contrast, task success declines for  
\texttt{LLaMA-3.3-70B} 
when the user simulator is too small or behaves inconsistently.

To quantify this effect, we introduce \textbf{\textsc{us}-spread}, a robustness metric defined as the range ($\text{maximum} -  \text{minimum}$) of task success rates achieved by a dialogue system when evaluated across different user simulator models. A lower \textsc{us}-spread indicates greater robustness against varying user behaviour. 
We observe that a monolithic dialogue system based on \verb|Qwen2.5-32B| achieves a spread of $0.58$ (see Table~\ref{tab:usvsdsperf}), while \verb|LLaMA-3.3-70B| has a slightly higher spread of $0.62$. 
Based on these findings, we recommend testing against a variety of user simulator models and using \textsc{us}-spread to assess robustness, during development.

\begin{table}[t]
  \begin{minipage}[t]{0.48\textwidth}
    \vspace{0pt} 
      \scriptsize    
    \begin{tabular}{lcccccc}
        \toprule
        \multirow{2}{*}{\textbf{Model (US)}} & \multicolumn{3}{c}{\textbf{Qwen2.5-32B (DS)}} & \multicolumn{3}{c}{\textbf{Llama-3.3-70B (DS)}} \\
        & \textbf{M} & \textbf{MP} & \textbf{ML} & \textbf{M} & \textbf{MP} & \textbf{ML} \\
        \midrule
          \verb|Llama-3.2-1B|   & 0.42 & 0.34 & 0.32 & 0.28 & 0.27 & 0.18 \\
          \verb|Llama-3.2-3B|   & 0.75 & 0.55 & 0.85 & 0.62 & 0.40 & 0.62 \\
          \verb|Qwen2.5-7B|     & 0.47 & 0.42 & 0.35 & 0.45 & 0.23 & 0.17 \\
          \verb|Llama-3.1-8B|   & 0.77 & 0.55 & 0.80 & 0.67 & 0.32 & 0.62 \\
          \verb|Qwen2.5-32B|    & 0.95 & 0.58 & 0.82 & 0.83 & 0.53 & 0.70 \\
          \verb|Llama-3.3-70B|  & \textbf{1.00} & \textbf{0.80} & \textbf{0.93} & 0.90 & 0.53 & 0.80 \\ \hline
          \verb|User Spread|    & 0.58 & \textbf{0.46} & 0.61 & 0.62 & 0.25 & 0.62 \\
        \bottomrule
      \end{tabular}
    \captionof{table}{Task success (booking) accuracy across different LLM pairings as user simulators (US) and dialogue systems (DS) in the \clemtodd\ framework. 
    Rows indicate the models used as US, and columns show the DS models evaluated in three architectural variants: Monolithic (M), Modular-Programmatic (MP), and Modular-LLM (ML). The bottom-most row (User Spread) reports the standard deviation across user simulators for each DS configuration, reflecting its sensitivity to variation in simulator behavior. Lower User Spread values indicate greater robustness, but should be considered alongside overall task success.
    } 
    \label{tab:usvsdsperf}
  \end{minipage}

\end{table}

While \verb|Llama-3.3-70B| achieves the highest task success rate overall as a user simulator, \verb|Qwen2.5-32B| closely matches its performance across both automatic and human evaluations, demonstrating strong task success, high naturalness, and competitive coherence. Given its significantly smaller model size, we select \verb|Qwen2.5-32B| as the user simulator for all subsequent experiments, balancing performance with computational efficiency.

\subsubsection{Evaluating Dialogue Systems}
With the user simulator model fixed, we can now systematically evaluate dialogue system realisation strategies, and the model realising them. \textit{How do different dialogue system configurations, varying in model size, prompting strategy, and architectural design, compare in terms of task success?}

The \clemtodd\ framework streamlines experimentation across models and configurations. We assess systems using a wide range of LLMs—spanning $1$B to $70$B parameters and model families such as Qwen, LLaMA, and GPT. Our objective is to identify  
how architectural choices impact performance, and to what extent model size influences outcomes. 

As mentioned in Section~\ref{sec:methodology}, we evaluated three different dialogue systems and the results are reported in Table~\ref{tab:archvsperf} based on standard metrics for task-oriented dialogue~\citep{DBLP:conf/emnlp/BudzianowskiWTC18}: \textit{Inform} (I), and \textit{Booking Accuracy} (B)~\citep{DBLP:conf/acl/XuMYSH24}, with \textit{Booking Accuracy} used as the primary indicator of end-to-end task success.

\begin{table*}
\scriptsize
\centering
    \begin{tabular}{lccccccccccc}
      \hline
      \multirow{2}{*}{\textbf{Model}} & \multicolumn{2}{c}{\textbf{Monolithic}} & \multicolumn{2}{c}{\textbf{Modular-Prog}} & \multicolumn{2}{c}{\textbf{Modular-LLM}} & \multicolumn{2}{c}{\textbf{\citet{DBLP:conf/acl/XuMYSH24}}} & \multicolumn{2}{c}{\textbf{\citet{DBLP:conf/sigdial/HudecekD23}}} \\
                     & Inform & Booking & Inform & Booking & Inform & Booking & Inform & Booking & Inform & Booking \\
      \hline
      \verb|Llama-3.2-1B|   & 0.00 & 0.00 & 0.00 & 0.00 & 0.00 & 0.00 & 0.00 & 0.00 & 0.20 & - \\
      \verb|Llama-3.2-3B|   & 0.05 & 0.05 & 0.0 & 0.0 & 0.02 & 0.02 & 0.0 & 0.0 & 0.20 & - \\
      \verb|Qwen2.5-7B|    & 0.09 & 0.09 & 0.28 & 0.26 & 0.25 & 0.24 & 0.32 & 0.30 & 0.25 & - \\
      \verb|Llama-3.1-8B|   & 0.20 & 0.18 & 0.0 & 0.0 & 0.08 & 0.06 & 0.12 & 0.09 & 0.28 & - \\
      \verb|Qwen2.5-32B|   & 0.76 & 0.71 & 0.44 & 0.41 & 0.71 & 0.68 & 0.63 & 0.60 & 0.52 & - \\
      \verb|Llama-3.3-70B|  & 0.70 & 0.69 & 0.43 & 0.43 & 0.53 & 0.52 & 0.12 & 0.12 & 0.32 & - \\
      \verb|GPT-4o|  & \textbf{0.82} & \textbf{0.81} & \textbf{0.65} & \textbf{0.65} & \textbf{0.85} & \textbf{0.84} & \textbf{0.75} & \textbf{0.73} & 0.42 & - \\      
      \hline
    \end{tabular}
  \caption{Performance of various dialogue system architectures within the \clemtodd\ framework. The table presents inform(I), and booking accuracy (B) for Monolithic, Modular-Prog, and Modular-LLM architectures across different LLMs. Systems based on \citet{DBLP:conf/acl/XuMYSH24} and \citet{DBLP:conf/sigdial/HudecekD23} are also evaluated within the \clemtodd\ setup. Larger models consistently yield better performance. \citet{DBLP:conf/sigdial/HudecekD23}'s system does not support booking accuracy computation. }
  \label{tab:archvsperf}
\end{table*}

At smaller scales ($1$B–$3$B), all architectures perform poorly (Booking Rate $\sim 0.05$ ), highlighting the limitations of small models for complex dialogue. We observe that a majority of conversations were aborted prematurely due to models failing to adhere to the expected response format constraints, leading to incomplete task execution and significantly lower task success rates.

At the other end of the scale, \verb|GPT-4o| achieves higher booking rates in modular configurations, reaching $\sim 0.84$. In monolithic setups, most failures occur in \textit{train domain} dialogues involving \textit{leave after} or \textit{arrival by} constraints, where the model often produces incorrect output formats (see Appendix~\ref{sec:response-format-issues}), leading to database failures and mismatches with ground truth goals. In contrast, the structured decomposition in modular systems helps the model reliably follow schema-specific outputs, improving constraint handling and task success.

In contrast, open-weight models perform better in monolithic configurations, achieving booking success rates around $\sim 0.77$, but show significantly reduced performance in modular setups. Most failures stem from invalid response formats or prematurely signalling the end of conversation without conversing about all the goals 
(see Appendix~\ref{sec:response-format-issues}). In modular configurations, the overall task is decomposed into subtasks, and individual modules operate without access to the full dialogue history. This limited context appears to hinder the model's ability to consistently satisfy goal constraints, reducing task success. In monolithic setups, a detailed analysis shows that these models can reach booking rates as high as $0.95$ (\verb|Qwen2.5–32B|) for single-domain dialogues (see Table~\ref{tab:domainwise_results} for the breakdown of results across domains). 
However, their performance drops in multi-domain scenarios due to increased complexity and cross-domain goal tracking.

\begin{figure}[t]
  \begin{minipage}[t]{0.48\textwidth}
    \vspace{0pt} 
    \scriptsize    
    \begin{tabular}{lcccccc}
      \hline
      \multirow{2}{*}{\textbf{Model}} & \multicolumn{2}{c}{\textbf{Monolithic}} & \multicolumn{2}{c}{\textbf{Modular-Prog}} & \multicolumn{2}{c}{\textbf{Modular-LLM}} \\
                     & T(\$) & F(\$) & T(\$) & F(\$) & T(\$) & F(\$) \\
      \hline
      \verb|Llama-3.2-1B|   & 0.009 & 0.007 & 0.009 & 0.007 & 0.0008 & 0.007 \\
      \verb|Llama-3.2-3B|   & 0.007 & 0.466 & 0.007 & 0.362 & 0.002 & 0.103 \\
      \verb|Qwen2.5-7B|    & 0.096 & 0.091 & 0.010 & 0.374 & 0.004 & 0.137 \\
      \verb|Llama-3.1-8B|   & 0.219 & 0.442 & 0.011 & 0.667 & 0.003 & 0.298 \\
      \verb|Qwen2.5-32B|   & 0.007 & 0.173 & 0.059 & 0.487 & 0.009 & 0.258 \\
      \verb|Llama-3.3-70B|  & 0.218 & 0.502 & 0.020 & 1.333 & 0.007 & 1.358 \\
      \verb|GPT-4o|  & 0.019 & NA & 2.113 & NA & 0.031 & NA \\
      \hline
    \end{tabular}
    \captionof{table}{Cost comparison of dialogue system architectures within the \clemtodd\ framework. The table reports token-based cost (T) and FLOPS cost (F), both measured in USD (\$) per dialogue, for Monolithic, Modular-Prog, and Modular-LLM architectures across different LLMs.} 
    \label{tab:costcomp}
  \end{minipage}
\end{figure}

\subsection{Evaluating Computational Efficiency of Dialogue Systems}
\textit{What are the computational cost patterns across dialogue system architectures and model scales?} 
We evaluated both token and FLOPs-based costs across all the three configurations using a standardized estimation approach~\citep{DBLP:journals/corr/abs-2001-08361, DBLP:journals/jmlr/ChowdheryNDBMRBCSGSSTMRBTSPRDHPBAI23}. More details on the cost computation are available in Appendix~\ref{sec:appendix-cost-estimation}

\begin{figure}[t]
  \begin{minipage}[t]{0.48\textwidth}
    \vspace{0pt} 
    \scriptsize    
    \begin{tabular}{lccc}
      \hline
      \textbf{Model} & \textbf{Monolithic} & \textbf{Modular-Prog} & \textbf{Modular-LLM} \\
      \hline
      \verb|Llama-3.2-1B|   & 18.85 & 1.61 & 18.72 \\
      \verb|Llama-3.2-3B|   & 5.21 & 8.32 & 16.56  \\
      \verb|Qwen2.5-7B|    & 6.69 & 11.38 & 13.70  \\
      \verb|Llama-3.1-8B|   & 5.97 & 11.16 & 42.08  \\
      \verb|Qwen2.5-32B|   & 8.52 & 10.16 & 7.17  \\
      \verb|Llama-3.3-70B|  & 25.33 & 34.07 & 47.98  \\
      \verb|GPT-4o|  & 4.10 & 8.61 & 4.37  \\
      \hline
    \end{tabular}
    \captionof{table}{Latency time (in seconds) comparison across dialogue system architectures in the \clemtodd\ framework. The table reports end-to-end response time for Monolithic, Modular-Prog, and Modular-LLM architectures evaluated across different LLMs.} 
    \label{tab:resptimecomp}
  \end{minipage}
\end{figure}

It is important to note that token costs are not solely determined by the number of tokens used. They also depend on the pricing set by API providers for each specific model. For example, some providers offer significantly lower input/output token prices for certain models, which can reduce the apparent cost even if token usage is high. Therefore, comparisons of token cost across architectures should be interpreted with this pricing variability in mind. While our cost estimates use OpenRouter pricing at the time of writing as a consistent baseline,\footnote{%
    \url{https://openrouter.ai}
}
these values may differ in production settings depending on the provider and deployment context.

As shown in Table~\ref{tab:costcomp}, 
the Monolithic architecture has the lowest computational cost due to its single-pass execution, avoiding repeated LLM calls. In contrast, Modular Prog and Modular LLM incur higher costs. 
Modular Prog has the highest FLOP cost, up to $\$1.33$ with \verb|Llama-3.3-70B|, $\sim 2$ times more than its Monolithic counterpart, reflecting poorly on efficiency. 

Additionally, we computed the latency time for each system, defined as the total time elapsed from the first interaction initiated by the user simulator until the task completion or the maximum number of dialogue turns ($15$) is reached. As shown in Table~\ref{tab:resptimecomp}, Monolithic systems exhibit the lowest latency times across most configurations. This is expected, as Monolithic systems involve a single model call per turn, whereas the Modular-Prog and Modular-LLM systems require multiple calls to sub-modules, increasing overall latency. The highest latency is observed for \texttt{Llama-3.3-70B}, which was executed on two GPUs, suggesting that model size and hardware configuration contribute significantly to response time. Among the modular variants, Modular-LLM systems generally exhibit higher latency than Modular-Prog systems, which can be attributed to the overhead of using an LLM as the dialogue manager in addition to the other modules. In contrast, the lowest latency is observed for \texttt{Llama-3.2-1B}, which is primarily due to early termination of many dialogues resulting from format violations leading to fewer turns and thus shorter overall latency.

\begin{figure}[t]
  \begin{minipage}[t]{0.49\textwidth}
    \vspace{0pt} 
    \includegraphics[width=\textwidth]{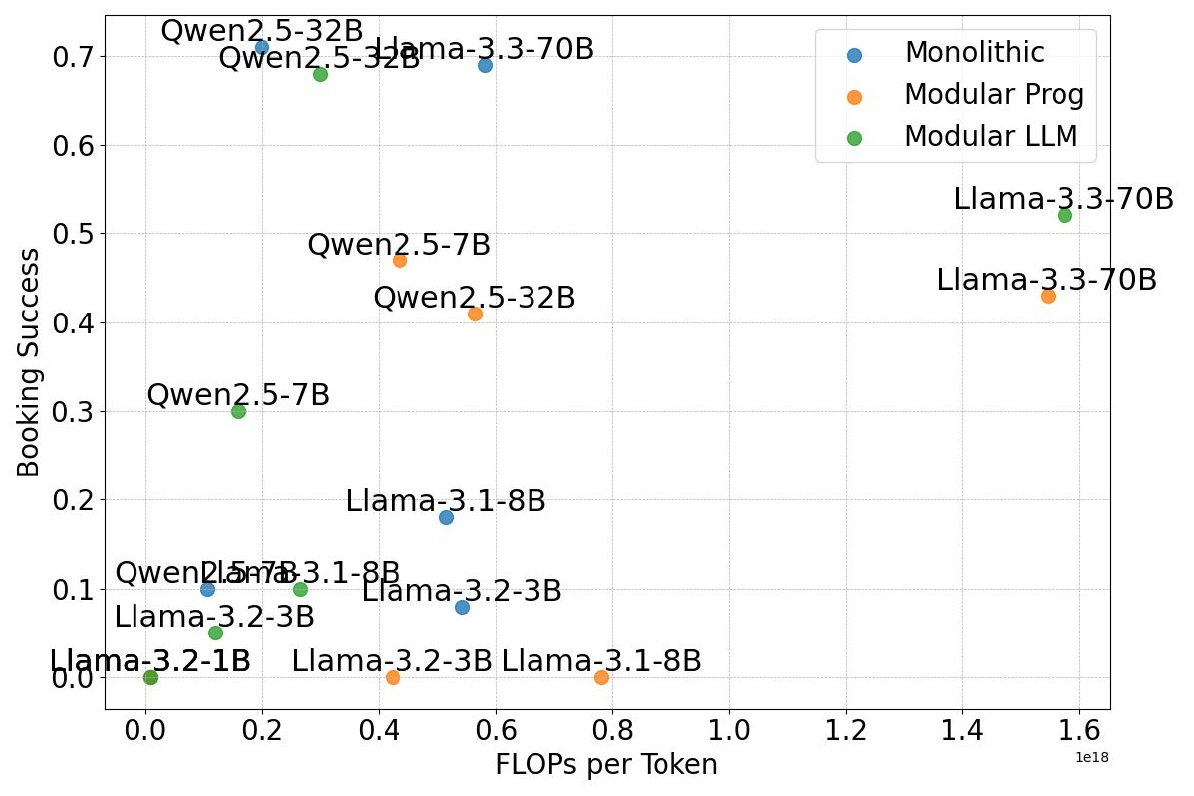}
    \captionof{figure}{Trade-off between computational cost and booking success across dialogue architectures.  The x-axis shows the FLOPs per token (computational cost), and the y-axis shows booking accuracy (task performance). Monolithic and Modular-LLM architectures tend to achieve higher booking accuracy at lower or comparable cost compared to Modular-Prog.
    }
    \label{fig:costvsperf}
  \end{minipage}
\end{figure}

These results underscore a key trade-off: modular architectures enhance interpretability and task decomposition but at a significant computational and token cost, especially at scale.

Further more, we want to analyse the trade-offs (see Figure~\ref{fig:costvsperf}) between model scale and computational cost (e.g., API usage, FLOPs). 
Larger models like \verb|Llama-3.3-70B| and \verb|Qwen2.5–32B| achieve markedly higher task performance in both monolithic and modular settings (e.g., \verb|Qwen2.5–32B| Monolithic: Booking Rate = $0.71$). Among variants, Modular-LLM strikes a favorable balance, delivering comparable performance (e.g., \verb|Qwen2.5–32B|: $0.68$ vs. $0.41$ in Mod-P) while reducing FLOP cost ($\$0.26$ vs. $\$0.49$). However, modular systems incur higher inference overhead due to multiple LLM calls per turn. This underscores a key trade-off: \textbf{monolithic systems offer inference efficiency but limited flexibility}, \textbf{whereas modular systems support extensibility and interpretability at greater computational cost}.

Overall, architecture selection should align with application-specific priorities. Monolithic setups are preferable when minimizing compute or API usage is critical. Modular architectures, particularly Modular-LLM, offer a compelling compromise.

\subsection{Evaluating the Adaptability of \clemtodd\ to New Domains}
\label{sec:clemtod-extendability}
\textit{Can \clemtodd\ be adapted to new domains and data configurations?} \clemtodd\ retains the design philosophy of the underlying ``self-play'' framework by separating the evaluation setup from the definition of the test instances. 
This allows evaluation 
with new instances (here, new goals), without any changes to the underlying 
logic. To assess this adaptability, we created two new synthetic datasets (see Appendix~\ref{sec:appendix-synthetic-data-generation}).

The first follows the MultiWOZ style but features unseen goal combinations that do not appear in the original dataset, while preserving task structure (see Appendix~\ref{sec:appendix-synthetic-data-generation}). The second contains intentionally unrealistic scenarios, such as \textit{booking a dragon for travel to London}, to test how systems handle inputs that are likely to differ from any data encountered during training.

We evaluated our proposed dialogue systems 
using the new datasets, while keeping the experimental configuration consistent with earlier evaluations. \texttt{Qwen2.5-32B} is used as the user simulator and we varied the dialogue system model. Table~\ref{tab:synthetic-multiwozstyle-overall_results} 
shows that models exhibit closer task success for the MultiWOZ-style data but struggled with unrealistic tasks.

\begin{figure}[t]
  \begin{minipage}[t]{0.48\textwidth}
    \vspace{0pt} 
    \scriptsize
    \begin{tabular}{lcccccc}
      \toprule
      \multirow{2}{*}{\textbf{Model}} & \multicolumn{3}{c}{\textbf{Multiwoz-Style}} & \multicolumn{3}{c}{\textbf{Unrealistic}} \\ & I & S & B & I & S & B \\
      \midrule
      \verb|Llama-3.2-1B|   & 0.00 & 0.00 & 0.00 & 0.00 & 0.00 & 0.00 \\
      \verb|Llama-3.2-3B|   & 0.18 & 0.08 & 0.18 & 0.03 & 0.02 & 0.02\\
      \verb|Qwen2.5-7B|    & 0.13 & 0.03 & 0.13 & 0.08 & 0.03 & 0.03\\
      \verb|Llama-3.1-8B|   & 0.26 & 0.08 & 0.26 & 0.08 & 0.03 & 0.07\\
      \verb|Qwen2.5-32B|   & 0.64 & 0.39 & \textbf{0.64} & 0.41 & 0.28 & 0.18\\
      \verb|Llama-3.3-70B|  & 0.54 & 0.24 & 0.54 & 0.20 & 0.11 & 0.07\\
      \verb|GPT-4o|  & 0.68 & 0.38 & \textbf{0.64} & 0.60 & 0.35 & \textbf{0.33}\\
      \bottomrule
    \end{tabular}
    \captionof{table}{Performance of monolithic dialogue systems evaluated for inform (I), success (S) and booking accuracy (B) with Qwen2.5-32B as the user simulator on the MultiWOZ-style synthetic and unrealistic datasets.} 
    \label{tab:synthetic-multiwozstyle-overall_results}
  \end{minipage}
\end{figure}

When the user simulator poses uncommon requests, such as booking a table for $-2$ people or \textit{reserving a stay in a dungeon in the middle of the ocean}, the dialogue systems often attempt to correct. For instance, it flags the negative party (-2) size as a likely typo, which the user simulator adjusts, resulting in a deviation from the original goal. Similarly, when handling unusual accommodation requests, the dialogue systems steering the user to consider more conventional hotel options. These interactions, while human-like, differ from the ground truth annotations, which leads to lower task success scores.
Among all models evaluated, \texttt{GPT-4o} and \texttt{Qwen2.5-32B} able to accommodate such unconventional requests to some extent.

Together, these results indicate that \clemtodd\ enables controlled evaluation on new data distributions and supports robust testing beyond fixed benchmarks. It provides a platform for analyzing how dialogue systems handle unseen goals, detect potential overfitting, and generalize to tasks outside their training exposure.

\section{Conclusion}
We propose \clemtodd\, a framework designed for the systematic evaluation of LLM-based task-oriented dialogue systems. By leveraging a structured self-play paradigm, \clemtodd\ enables consistent and modular assessment across diverse dialogue system architectures and user simulators. Our experiments on standard (MultiWOZ) benchmarks, demonstrate that while large-scale monolithic models offer better performance at low computational cost, modular architectures, especially Modular-LLM variants, achieve reasonable trade-offs between performance and efficiency. Furthermore, we show that \clemtodd\ can be adapted to evaluate dialogue systems in unseen and unrealistic domains beyond fixed datasets.

\section*{Limitations}
Although \clemtodd\ offers flexibility and adaptability, it has limitations that call for future research. First, the design choice of using a simple, classical pipeline in our proposed modular dialogue systems was intended to assess LLM performance in a controlled setting. However, the relatively lower scores (for modular-program variant) may not reflect model limitations alone, but also limitations of the pipeline design itself, which may be too constrained to capture more complex dialogue behaviors. Second, our proposed setup assumes strict adherence to response formats (as defined by the Tool Schema), which smaller models frequently violate. This leads to premature termination of dialogues and may underestimate the true capabilities of such models. Third, our current evaluation includes only GPT-4o among closed-weight models, limiting broader comparison with other proprietary LLMs (Claude, Gemini etc.). Fourth, while the current self-play setup is designed to be extensible, it is limited to two-player interactions. Extending the framework to support multi-agent dialogue scenarios is an open direction for future exploration.

\section*{Acknowledgments}
The work reported here has been funded by the Bundesministerium für Bildung und Forschung (BMBF, German Federal Ministry of Research), project "COCOBOTS" (01IS21102A) and Deutsche Forschungsgemeinschaft (DFG, German Research Foundation) grant 423217434 (“RECOLAGE”). We thank the anonymous reviewers for their helpful feedback.

\bibliography{main}

\begin{thebibliography}{45}
\providecommand{\natexlab}[1]{#1}

\bibitem[{Balakrishnan et~al.(2019)Balakrishnan, Rao, Upasani, White, and Subba}]{DBLP:conf/acl/BalakrishnanRUW19}
Anusha Balakrishnan, Jinfeng Rao, Kartikeya Upasani, Michael White, and Rajen Subba. 2019.
\newblock \href {https://doi.org/10.18653/V1/P19-1080} {Constrained decoding for neural {NLG} from compositional representations in task-oriented dialogue}.
\newblock In \emph{Proceedings of the 57th Conference of the Association for Computational Linguistics, {ACL} 2019, Florence, Italy, July 28- August 2, 2019, Volume 1: Long Papers}, pages 831--844. Association for Computational Linguistics.

\bibitem[{Bavaresco et~al.(2024)Bavaresco, Bernardi, Bertolazzi, Elliott, Fernández, Gatt, Ghaleb, Giulianelli, Hanna, Koller, Martins, Mondorf, Neplenbroek, Pezzelle, Plank, Schlangen, Suglia, Surikuchi, Takmaz, and Testoni}]{bavaresco2024llmsinsteadhumanjudges}
Anna Bavaresco, Raffaella Bernardi, Leonardo Bertolazzi, Desmond Elliott, Raquel Fernández, Albert Gatt, Esam Ghaleb, Mario Giulianelli, Michael Hanna, Alexander Koller, André F.~T. Martins, Philipp Mondorf, Vera Neplenbroek, Sandro Pezzelle, Barbara Plank, David Schlangen, Alessandro Suglia, Aditya~K Surikuchi, Ece Takmaz, and Alberto Testoni. 2024.
\newblock \href {https://arxiv.org/abs/2406.18403} {Llms instead of human judges? a large scale empirical study across 20 nlp evaluation tasks}.
\newblock \emph{Preprint}, arXiv:2406.18403.

\bibitem[{Braggaar et~al.(2023)Braggaar, Liebrecht, van Miltenburg, and Krahmer}]{DBLP:journals/corr/abs-2312-13871}
Anouck Braggaar, Christine Liebrecht, Emiel van Miltenburg, and Emiel~J. Krahmer. 2023.
\newblock \href {https://doi.org/10.48550/ARXIV.2312.13871} {Evaluating task-oriented dialogue systems: {A} systematic review of measures, constructs and their operationalisations}.
\newblock \emph{CoRR}, abs/2312.13871.

\bibitem[{Brown et~al.(2020)Brown, Mann, Ryder, Subbiah, Kaplan, Dhariwal, Neelakantan, Shyam, Sastry, Askell, Agarwal, Herbert{-}Voss, Krueger, Henighan, Child, Ramesh, Ziegler, Wu, Winter, Hesse, Chen, Sigler, Litwin, Gray, Chess, Clark, Berner, McCandlish, Radford, Sutskever, and Amodei}]{DBLP:conf/nips/BrownMRSKDNSSAA20}
Tom~B. Brown, Benjamin Mann, Nick Ryder, Melanie Subbiah, Jared Kaplan, Prafulla Dhariwal, Arvind Neelakantan, Pranav Shyam, Girish Sastry, Amanda Askell, Sandhini Agarwal, Ariel Herbert{-}Voss, Gretchen Krueger, Tom Henighan, Rewon Child, Aditya Ramesh, Daniel~M. Ziegler, Jeffrey Wu, Clemens Winter, and 12 others. 2020.
\newblock \href {https://proceedings.neurips.cc/paper/2020/hash/1457c0d6bfcb4967418bfb8ac142f64a-Abstract.html} {Language models are few-shot learners}.
\newblock In \emph{Advances in Neural Information Processing Systems 33: Annual Conference on Neural Information Processing Systems 2020, NeurIPS 2020, December 6-12, 2020, virtual}.

\bibitem[{Budzianowski et~al.(2018)Budzianowski, Wen, Tseng, Casanueva, Ultes, Ramadan, and Gasic}]{DBLP:conf/emnlp/BudzianowskiWTC18}
Pawel Budzianowski, Tsung{-}Hsien Wen, Bo{-}Hsiang Tseng, I{\~{n}}igo Casanueva, Stefan Ultes, Osman Ramadan, and Milica Gasic. 2018.
\newblock \href {https://aclanthology.org/D18-1547/} {Multiwoz - {A} large-scale multi-domain wizard-of-oz dataset for task-oriented dialogue modelling}.
\newblock In \emph{Proceedings of the 2018 Conference on Empirical Methods in Natural Language Processing, Brussels, Belgium, October 31 - November 4, 2018}, pages 5016--5026. Association for Computational Linguistics.

\bibitem[{Casas et~al.(2020)Casas, Tricot, Khaled, Mugellini, and Cudr{\'{e}}{-}Mauroux}]{DBLP:conf/icmi/CasasTKMC20}
Jacky Casas, Marc{-}Olivier Tricot, Omar~Abou Khaled, Elena Mugellini, and Philippe Cudr{\'{e}}{-}Mauroux. 2020.
\newblock \href {https://doi.org/10.1145/3395035.3425319} {Trends {\&} methods in chatbot evaluation}.
\newblock In \emph{Companion Publication of the 2020 International Conference on Multimodal Interaction, {ICMI} Companion 2020, Virtual Event, The Netherlands, October, 2020}, pages 280--286. {ACM}.

\bibitem[{Chalamalasetti et~al.(2023)Chalamalasetti, G{\"{o}}tze, Hakimov, Madureira, Sadler, and Schlangen}]{DBLP:conf/emnlp/ChalamalasettiG23}
Kranti Chalamalasetti, Jana G{\"{o}}tze, Sherzod Hakimov, Brielen Madureira, Philipp Sadler, and David Schlangen. 2023.
\newblock \href {https://doi.org/10.18653/V1/2023.EMNLP-MAIN.689} {clembench: Using game play to evaluate chat-optimized language models as conversational agents}.
\newblock In \emph{Proceedings of the 2023 Conference on Empirical Methods in Natural Language Processing, {EMNLP} 2023, Singapore, December 6-10, 2023}, pages 11174--11219. Association for Computational Linguistics.

\bibitem[{Chan et~al.(2024)Chan, Chen, Su, Yu, Xue, Zhang, Fu, and Liu}]{DBLP:conf/iclr/ChanCSYXZF024}
Chi{-}Min Chan, Weize Chen, Yusheng Su, Jianxuan Yu, Wei Xue, Shanghang Zhang, Jie Fu, and Zhiyuan Liu. 2024.
\newblock \href {https://openreview.net/forum?id=FQepisCUWu} {Chateval: Towards better llm-based evaluators through multi-agent debate}.
\newblock In \emph{The Twelfth International Conference on Learning Representations, {ICLR} 2024, Vienna, Austria, May 7-11, 2024}. OpenReview.net.

\bibitem[{Chen et~al.(2019)Chen, Xu, and Xu}]{chen-etal-2019-working}
Xiuyi Chen, Jiaming Xu, and Bo~Xu. 2019.
\newblock \href {https://doi.org/10.18653/v1/P19-1258} {A working memory model for task-oriented dialog response generation}.
\newblock In \emph{Proceedings of the 57th Annual Meeting of the Association for Computational Linguistics}, pages 2687--2693, Florence, Italy. Association for Computational Linguistics.

\bibitem[{Cheng et~al.(2024)Cheng, Hu, Xu, Zhang, Dai, Han, Du, and Li}]{DBLP:conf/nips/ChengHXZDHDL24}
Pengyu Cheng, Tianhao Hu, Han Xu, Zhisong Zhang, Yong Dai, Lei Han, Nan Du, and Xiaolong Li. 2024.
\newblock \href {http://papers.nips.cc/paper\_files/paper/2024/hash/e4be7e9867ef163563f4a5e90cec478f-Abstract-Conference.html} {Self-playing adversarial language game enhances {LLM} reasoning}.
\newblock In \emph{Advances in Neural Information Processing Systems 38: Annual Conference on Neural Information Processing Systems 2024, NeurIPS 2024, Vancouver, BC, Canada, December 10 - 15, 2024}.

\bibitem[{Cheng et~al.(2022)Cheng, Li, Quan, Gao, Mou, and Qiu}]{cheng-etal-2022-multiwoz}
Qinyuan Cheng, Linyang Li, Guofeng Quan, Feng Gao, Xiaofeng Mou, and Xipeng Qiu. 2022.
\newblock \href {https://doi.org/10.18653/v1/2022.findings-emnlp.90} {Is {M}ulti{WOZ} a solved task? an interactive {TOD} evaluation framework with user simulator}.
\newblock In \emph{Findings of the Association for Computational Linguistics: EMNLP 2022}, pages 1248--1259, Abu Dhabi, United Arab Emirates. Association for Computational Linguistics.

\bibitem[{Chowdhery et~al.(2023)Chowdhery, Narang, Devlin, Bosma, Mishra, Roberts, Barham, Chung, Sutton, Gehrmann, Schuh, Shi, Tsvyashchenko, Maynez, Rao, Barnes, Tay, Shazeer, Prabhakaran, Reif, Du, Hutchinson, Pope, Bradbury, Austin, Isard, Gur{-}Ari, Yin, Duke, Levskaya, Ghemawat, Dev, Michalewski, Garcia, Misra, Robinson, Fedus, Zhou, Ippolito, Luan, Lim, Zoph, Spiridonov, Sepassi, Dohan, Agrawal, Omernick, Dai, Pillai, Pellat, Lewkowycz, Moreira, Child, Polozov, Lee, Zhou, Wang, Saeta, Diaz, Firat, Catasta, Wei, Meier{-}Hellstern, Eck, Dean, Petrov, and Fiedel}]{DBLP:journals/jmlr/ChowdheryNDBMRBCSGSSTMRBTSPRDHPBAI23}
Aakanksha Chowdhery, Sharan Narang, Jacob Devlin, Maarten Bosma, Gaurav Mishra, Adam Roberts, Paul Barham, Hyung~Won Chung, Charles Sutton, Sebastian Gehrmann, Parker Schuh, Kensen Shi, Sasha Tsvyashchenko, Joshua Maynez, Abhishek Rao, Parker Barnes, Yi~Tay, Noam Shazeer, Vinodkumar Prabhakaran, and 48 others. 2023.
\newblock \href {https://jmlr.org/papers/v24/22-1144.html} {Palm: Scaling language modeling with pathways}.
\newblock \emph{J. Mach. Learn. Res.}, 24:240:1--240:113.

\bibitem[{Chung et~al.(2023)Chung, Cahyawijaya, Wilie, Lovenia, and Fung}]{chung-etal-2023-instructtods}
Willy Chung, Samuel Cahyawijaya, Bryan Wilie, Holy Lovenia, and Pascale Fung. 2023.
\newblock \href {https://doi.org/10.18653/v1/2023.nlint-1.1} {{I}nstruct{TODS}: Large language models for end-to-end task-oriented dialogue systems}.
\newblock In \emph{Proceedings of the Second Workshop on Natural Language Interfaces}, pages 1--21. Association for Computational Linguistics.

\bibitem[{Davidson et~al.(2023)Davidson, Romeo, Shu, Gung, Gupta, Mansour, and Zhang}]{DBLP:journals/corr/abs-2309-13233}
Sam Davidson, Salvatore Romeo, Raphael Shu, James Gung, Arshit Gupta, Saab Mansour, and Yi~Zhang. 2023.
\newblock \href {https://doi.org/10.48550/ARXIV.2309.13233} {User simulation with large language models for evaluating task-oriented dialogue}.
\newblock \emph{CoRR}, abs/2309.13233.

\bibitem[{Dong et~al.(2025)Dong, Chen, and Yang}]{DBLP:conf/coling/DongCY25}
Wenjie Dong, Sirong Chen, and Yan Yang. 2025.
\newblock \href {https://aclanthology.org/2025.coling-main.614/} {Protod: Proactive task-oriented dialogue system based on large language model}.
\newblock In \emph{Proceedings of the 31st International Conference on Computational Linguistics, {COLING} 2025, Abu Dhabi, UAE, January 19-24, 2025}, pages 9147--9164. Association for Computational Linguistics.

\bibitem[{Elder et~al.(2020)Elder, O{'}Connor, and Foster}]{elder-etal-2020-make}
Henry Elder, Alexander O{'}Connor, and Jennifer Foster. 2020.
\newblock \href {https://doi.org/10.18653/v1/2020.emnlp-main.230} {How to make neural natural language generation as reliable as templates in task-oriented dialogue}.
\newblock In \emph{Proceedings of the 2020 Conference on Empirical Methods in Natural Language Processing (EMNLP)}, pages 2877--2888, Online. Association for Computational Linguistics.

\bibitem[{Eric et~al.(2017)Eric, Krishnan, Charette, and Manning}]{eric-etal-2017-key}
Mihail Eric, Lakshmi Krishnan, Francois Charette, and Christopher~D. Manning. 2017.
\newblock \href {https://doi.org/10.18653/v1/W17-5506} {Key-value retrieval networks for task-oriented dialogue}.
\newblock In \emph{Proceedings of the 18th Annual {SIG}dial Meeting on Discourse and Dialogue}, pages 37--49, Saarbr{\"u}cken, Germany. Association for Computational Linguistics.

\bibitem[{Ghandeharioun et~al.(2019)Ghandeharioun, Shen, Jaques, Ferguson, Jones, Lapedriza, and Picard}]{10.5555/3454287.3455511}
Asma Ghandeharioun, Judy~Hanwen Shen, Natasha Jaques, Craig Ferguson, Noah Jones, Agata Lapedriza, and Rosalind Picard. 2019.
\newblock \emph{Approximating interactive human evaluation with self-play for open-domain dialog systems}.
\newblock Curran Associates Inc., Red Hook, NY, USA.

\bibitem[{Grattafiori et~al.(2024)Grattafiori, Dubey, Jauhri, Pandey, Kadian, and et~al.}]{llama31}
Aaron Grattafiori, Abhimanyu Dubey, Abhinav Jauhri, Abhinav Pandey, Abhishek Kadian, and et~al. 2024.
\newblock \href {https://arxiv.org/abs/2407.21783} {The llama 3 herd of models}.
\newblock \emph{Preprint}, arXiv:2407.21783.

\bibitem[{Guertler et~al.(2025)Guertler, Cheng, Yu, Liu, Choshen, and Tan}]{guertler2025textarena}
Leon Guertler, Bobby Cheng, Simon Yu, Bo~Liu, Leshem Choshen, and Cheston Tan. 2025.
\newblock Textarena.
\newblock \emph{arXiv preprint arXiv:2504.11442}.

\bibitem[{Gupta et~al.(2024)Gupta, Ravichandran, Zhang, Shah, Beniwal, and Sadagopan}]{DBLP:journals/corr/abs-2411-00427}
Aman Gupta, Anirudh Ravichandran, Ziji Zhang, Swair Shah, Anurag Beniwal, and Narayanan Sadagopan. 2024.
\newblock \href {https://doi.org/10.48550/ARXIV.2411.00427} {{DARD:} {A} multi-agent approach for task-oriented dialog systems}.
\newblock \emph{CoRR}, abs/2411.00427.

\bibitem[{Hashemi et~al.(2024)Hashemi, Eisner, Rosset, Durme, and Kedzie}]{DBLP:conf/acl/HashemiERDK24}
Helia Hashemi, Jason Eisner, Corby Rosset, Benjamin~Van Durme, and Chris Kedzie. 2024.
\newblock \href {https://doi.org/10.18653/V1/2024.ACL-LONG.745} {Llm-rubric: {A} multidimensional, calibrated approach to automated evaluation of natural language texts}.
\newblock In \emph{Proceedings of the 62nd Annual Meeting of the Association for Computational Linguistics (Volume 1: Long Papers), {ACL} 2024, Bangkok, Thailand, August 11-16, 2024}, pages 13806--13834. Association for Computational Linguistics.

\bibitem[{Hudecek and Dusek(2023)}]{DBLP:conf/sigdial/HudecekD23}
Vojtech Hudecek and Ondrej Dusek. 2023.
\newblock \href {https://doi.org/10.18653/V1/2023.SIGDIAL-1.21} {Are large language models all you need for task-oriented dialogue?}
\newblock In \emph{Proceedings of the 24th Meeting of the Special Interest Group on Discourse and Dialogue, {SIGDIAL} 2023, Prague, Czechia, September 11 - 15, 2023}, pages 216--228. Association for Computational Linguistics.

\bibitem[{Hung et~al.(2022)Hung, Lauscher, Vulic, Ponzetto, and Glavas}]{DBLP:conf/naacl/HungLVPG22}
Chia{-}Chien Hung, Anne Lauscher, Ivan Vulic, Simone~Paolo Ponzetto, and Goran Glavas. 2022.
\newblock \href {https://doi.org/10.18653/V1/2022.NAACL-MAIN.270} {Multi2woz: {A} robust multilingual dataset and conversational pretraining for task-oriented dialog}.
\newblock In \emph{Proceedings of the 2022 Conference of the North American Chapter of the Association for Computational Linguistics: Human Language Technologies, {NAACL} 2022, Seattle, WA, United States, July 10-15, 2022}, pages 3687--3703. Association for Computational Linguistics.

\bibitem[{Jia et~al.(2024)Jia, Komma, Leffel, Peng, Nagesh, Soliman, Galstyan, and Kumar}]{DBLP:conf/naacl/JiaKLPNSGK24}
Jinghan Jia, Abi Komma, Timothy Leffel, Xujun Peng, Ajay Nagesh, Tamer Soliman, Aram Galstyan, and Anoop Kumar. 2024.
\newblock \href {https://doi.org/10.18653/V1/2024.NAACL-INDUSTRY.30} {Leveraging llms for dialogue quality measurement}.
\newblock In \emph{Proceedings of the 2024 Conference of the North American Chapter of the Association for Computational Linguistics: Human Language Technologies: Industry Track, {NAACL} 2024, Mexico City, Mexico, June 16-21, 2024}, pages 359--367. Association for Computational Linguistics.

\bibitem[{Kaplan et~al.(2020)Kaplan, McCandlish, Henighan, Brown, Chess, Child, Gray, Radford, Wu, and Amodei}]{DBLP:journals/corr/abs-2001-08361}
Jared Kaplan, Sam McCandlish, Tom Henighan, Tom~B. Brown, Benjamin Chess, Rewon Child, Scott Gray, Alec Radford, Jeffrey Wu, and Dario Amodei. 2020.
\newblock \href {https://arxiv.org/abs/2001.08361} {Scaling laws for neural language models}.
\newblock \emph{CoRR}, abs/2001.08361.

\bibitem[{Kazi et~al.(2024)Kazi, Lyu, Zhou, Hakkani{-}T{\"{u}}r, and Tur}]{DBLP:conf/slt/KaziLZHT24}
Taaha Kazi, Ruiliang Lyu, Sizhe Zhou, Dilek Hakkani{-}T{\"{u}}r, and Gokhan Tur. 2024.
\newblock \href {https://doi.org/10.1109/SLT61566.2024.10832298} {Large language models as user-agents for evaluating task-oriented-dialogue systems}.
\newblock In \emph{{IEEE} Spoken Language Technology Workshop, {SLT} 2024, Macao, December 2-5, 2024}, pages 913--920. {IEEE}.

\bibitem[{Kim et~al.(2024)Kim, Suk, Longpre, Lin, Shin, Welleck, Neubig, Lee, Lee, and Seo}]{DBLP:conf/emnlp/KimSLLSWNL0S24}
Seungone Kim, Juyoung Suk, Shayne Longpre, Bill~Yuchen Lin, Jamin Shin, Sean Welleck, Graham Neubig, Moontae Lee, Kyungjae Lee, and Minjoon Seo. 2024.
\newblock \href {https://aclanthology.org/2024.emnlp-main.248} {Prometheus 2: An open source language model specialized in evaluating other language models}.
\newblock In \emph{Proceedings of the 2024 Conference on Empirical Methods in Natural Language Processing, {EMNLP} 2024, Miami, FL, USA, November 12-16, 2024}, pages 4334--4353. Association for Computational Linguistics.

\bibitem[{Kim et~al.(2023)Kim, Shin, Kim, Bae, and Kim}]{DBLP:journals/corr/abs-2305-13857}
Takyoung Kim, Jamin Shin, Young{-}Ho Kim, Sanghwan Bae, and Sungdong Kim. 2023.
\newblock \href {https://doi.org/10.48550/ARXIV.2305.13857} {Revealing user familiarity bias in task-oriented dialogue via interactive evaluation}.
\newblock \emph{CoRR}, abs/2305.13857.

\bibitem[{Kojima et~al.(2022)Kojima, Gu, Reid, Matsuo, and Iwasawa}]{DBLP:conf/nips/KojimaGRMI22}
Takeshi Kojima, Shixiang~Shane Gu, Machel Reid, Yutaka Matsuo, and Yusuke Iwasawa. 2022.
\newblock \href {http://papers.nips.cc/paper\_files/paper/2022/hash/8bb0d291acd4acf06ef112099c16f326-Abstract-Conference.html} {Large language models are zero-shot reasoners}.
\newblock In \emph{Advances in Neural Information Processing Systems 35: Annual Conference on Neural Information Processing Systems 2022, NeurIPS 2022, New Orleans, LA, USA, November 28 - December 9, 2022}.

\bibitem[{Laidlaw et~al.(2024)Laidlaw, Bronstein, Guo, Feng, Berglund, Svegliato, Russell, and Dragan}]{laidlaw2024scalably}
Cassidy Laidlaw, Eli Bronstein, Timothy Guo, Dylan Feng, Lukas Berglund, Justin Svegliato, Stuart Russell, and Anca Dragan. 2024.
\newblock Scalably solving assistance games.
\newblock In \emph{ICML 2024 Workshop on Models of Human Feedback for AI Alignment}.

\bibitem[{Li et~al.(2017)Li, Chen, Li, Gao, and Celikyilmaz}]{DBLP:conf/ijcnlp/LiCLGC17}
Xiujun Li, Yun{-}Nung Chen, Lihong Li, Jianfeng Gao, and Asli Celikyilmaz. 2017.
\newblock \href {https://aclanthology.org/I17-1074/} {End-to-end task-completion neural dialogue systems}.
\newblock In \emph{Proceedings of the Eighth International Joint Conference on Natural Language Processing, {IJCNLP} 2017, Taipei, Taiwan, November 27 - December 1, 2017 - Volume 1: Long Papers}, pages 733--743. Asian Federation of Natural Language Processing.

\bibitem[{Liu et~al.(2023{\natexlab{a}})Liu, Yuan, Fu, Jiang, Hayashi, and Neubig}]{DBLP:journals/csur/LiuYFJHN23}
Pengfei Liu, Weizhe Yuan, Jinlan Fu, Zhengbao Jiang, Hiroaki Hayashi, and Graham Neubig. 2023{\natexlab{a}}.
\newblock \href {https://doi.org/10.1145/3560815} {Pre-train, prompt, and predict: {A} systematic survey of prompting methods in natural language processing}.
\newblock \emph{{ACM} Comput. Surv.}, 55(9):195:1--195:35.

\bibitem[{Liu et~al.(2023{\natexlab{b}})Liu, Iter, Xu, Wang, Xu, and Zhu}]{DBLP:conf/emnlp/LiuIXWXZ23}
Yang Liu, Dan Iter, Yichong Xu, Shuohang Wang, Ruochen Xu, and Chenguang Zhu. 2023{\natexlab{b}}.
\newblock \href {https://doi.org/10.18653/V1/2023.EMNLP-MAIN.153} {G-eval: {NLG} evaluation using gpt-4 with better human alignment}.
\newblock In \emph{Proceedings of the 2023 Conference on Empirical Methods in Natural Language Processing, {EMNLP} 2023, Singapore, December 6-10, 2023}, pages 2511--2522. Association for Computational Linguistics.

\bibitem[{McTear(2002)}]{McTear2002}
Michael~F McTear. 2002.
\newblock \href {https://doi.org/10.1145/505282.505285} {Spoken dialogue technology: enabling the conversational user interface}.
\newblock \emph{ACM Computing Surveys}, 34(1):90--169.
\newblock Publisher: School of Information and Software Engineering, University of Ulster.

\bibitem[{Pietquin and Hastie(2013)}]{DBLP:journals/ker/PietquinH13}
Olivier Pietquin and Helen~F. Hastie. 2013.
\newblock \href {https://doi.org/10.1017/S0269888912000343} {A survey on metrics for the evaluation of user simulations}.
\newblock \emph{Knowl. Eng. Rev.}, 28(1):59--73.

\bibitem[{Qiao et~al.(2023)Qiao, Wu, Liang, Li, and Duan}]{DBLP:journals/corr/abs-2308-10032}
Dan Qiao, Chenfei Wu, Yaobo Liang, Juntao Li, and Nan Duan. 2023.
\newblock \href {https://doi.org/10.48550/ARXIV.2308.10032} {Gameeval: Evaluating llms on conversational games}.
\newblock \emph{CoRR}, abs/2308.10032.

\bibitem[{Qwen et~al.(2025)Qwen, Yang, Yang, Zhang, Hui, and et~al.}]{qwen25}
Qwen, An~Yang, Baosong Yang, Beichen Zhang, Binyuan Hui, and et~al. 2025.
\newblock \href {https://arxiv.org/abs/2412.15115} {Qwen2.5 technical report}.
\newblock \emph{Preprint}, arXiv:2412.15115.

\bibitem[{Sekulic et~al.(2024)Sekulic, Terragni, Guimar{\~{a}}es, Khau, Guedes, Filipavicius, Manso, and Mathis}]{DBLP:journals/corr/abs-2402-13374}
Ivan Sekulic, Silvia Terragni, Victor Guimar{\~{a}}es, Nghia Khau, Bruna Guedes, Modestas Filipavicius, Andr{\'{e}}~Ferreira Manso, and Roland Mathis. 2024.
\newblock \href {https://doi.org/10.48550/ARXIV.2402.13374} {Reliable llm-based user simulator for task-oriented dialogue systems}.
\newblock \emph{CoRR}, abs/2402.13374.

\bibitem[{Thakur et~al.(2024)Thakur, Choudhary, Ramayapally, Vaidyanathan, and Hupkes}]{DBLP:journals/corr/abs-2406-12624}
Aman~Singh Thakur, Kartik Choudhary, Venkat~Srinik Ramayapally, Sankaran Vaidyanathan, and Dieuwke Hupkes. 2024.
\newblock \href {https://doi.org/10.48550/ARXIV.2406.12624} {Judging the judges: Evaluating alignment and vulnerabilities in llms-as-judges}.
\newblock \emph{CoRR}, abs/2406.12624.

\bibitem[{Wei et~al.(2022)Wei, Wang, Schuurmans, Bosma, Ichter, Xia, Chi, Le, and Zhou}]{DBLP:conf/nips/Wei0SBIXCLZ22}
Jason Wei, Xuezhi Wang, Dale Schuurmans, Maarten Bosma, Brian Ichter, Fei Xia, Ed~H. Chi, Quoc~V. Le, and Denny Zhou. 2022.
\newblock \href {http://papers.nips.cc/paper\_files/paper/2022/hash/9d5609613524ecf4f15af0f7b31abca4-Abstract-Conference.html} {Chain-of-thought prompting elicits reasoning in large language models}.
\newblock In \emph{Advances in Neural Information Processing Systems 35: Annual Conference on Neural Information Processing Systems 2022, NeurIPS 2022, New Orleans, LA, USA, November 28 - December 9, 2022}.

\bibitem[{Xu et~al.(2024)Xu, Mao, Yang, Sun, and Huang}]{DBLP:conf/acl/XuMYSH24}
Heng{-}Da Xu, Xian{-}Ling Mao, Puhai Yang, Fanshu Sun, and Heyan Huang. 2024.
\newblock \href {https://doi.org/10.18653/V1/2024.ACL-LONG.152} {Rethinking task-oriented dialogue systems: From complex modularity to zero-shot autonomous agent}.
\newblock In \emph{Proceedings of the 62nd Annual Meeting of the Association for Computational Linguistics (Volume 1: Long Papers), {ACL} 2024, Bangkok, Thailand, August 11-16, 2024}, pages 2748--2763. Association for Computational Linguistics.

\bibitem[{Yeh et~al.(2021)Yeh, Esk{\'{e}}nazi, and Mehri}]{DBLP:journals/corr/abs-2106-03706}
Yi{-}Ting Yeh, Maxine Esk{\'{e}}nazi, and Shikib Mehri. 2021.
\newblock \href {https://arxiv.org/abs/2106.03706} {A comprehensive assessment of dialog evaluation metrics}.
\newblock \emph{CoRR}, abs/2106.03706.

\bibitem[{Zheng et~al.(2023)Zheng, Chiang, Sheng, Zhuang, Wu, Zhuang, Lin, Li, Li, Xing, Zhang, Gonzalez, and Stoica}]{DBLP:conf/nips/ZhengC00WZL0LXZ23}
Lianmin Zheng, Wei{-}Lin Chiang, Ying Sheng, Siyuan Zhuang, Zhanghao Wu, Yonghao Zhuang, Zi~Lin, Zhuohan Li, Dacheng Li, Eric~P. Xing, Hao Zhang, Joseph~E. Gonzalez, and Ion Stoica. 2023.
\newblock \href {http://papers.nips.cc/paper\_files/paper/2023/hash/91f18a1287b398d378ef22505bf41832-Abstract-Datasets\_and\_Benchmarks.html} {Judging llm-as-a-judge with mt-bench and chatbot arena}.
\newblock In \emph{Advances in Neural Information Processing Systems 36: Annual Conference on Neural Information Processing Systems 2023, NeurIPS 2023, New Orleans, LA, USA, December 10 - 16, 2023}.

\bibitem[{Zhu et~al.(2020)Zhu, Zhang, Fang, Li, Takanobu, Li, Peng, Gao, Zhu, and Huang}]{DBLP:conf/acl/ZhuZFLTLPGZH20}
Qi~Zhu, Zheng Zhang, Yan Fang, Xiang Li, Ryuichi Takanobu, Jinchao Li, Baolin Peng, Jianfeng Gao, Xiaoyan Zhu, and Minlie Huang. 2020.
\newblock \href {https://doi.org/10.18653/V1/2020.ACL-DEMOS.19} {Convlab-2: An open-source toolkit for building, evaluating, and diagnosing dialogue systems}.
\newblock In \emph{Proceedings of the 58th Annual Meeting of the Association for Computational Linguistics: System Demonstrations, {ACL} 2020, Online, July 5-10, 2020}, pages 142--149. Association for Computational Linguistics.

\end{thebibliography}

\appendix

\section{Appendix}
\label{sec:appendix}

\subsection {Prompt Templates}
\label{sec:appendix-prompt-templates}
In the proposed tasks, LLMs are used in a variety of roles across different dialogue system architectures. Specifically, LLMs serve as user simulator (see Figure~\ref{fig:usimulator_prompt}), as a complete dialogue system (in the monolithic architecture) (see Figure~\ref{fig:monods_prompt}), as a dialogue manager (in modular-LLM architecture) (see Figure~\ref{fig:modllm_ds_prompt}), and as a individual dialogue modules such as intent detector (see Figure~\ref{fig:modprog_intent_detection}), slot extractor (see Figure~\ref{fig:modprog_slot_extraction}), and response generator (see Figure~\ref{fig:modprog_response_generation}) in both modular-LLM and modular-prog architectures.

Following standard prompting approaches~\citep{DBLP:conf/nips/BrownMRSKDNSSAA20, DBLP:conf/nips/Wei0SBIXCLZ22, DBLP:journals/csur/LiuYFJHN23}, all prompts were constructed in a zero-shot setting without the use of explicit  examples. Prompts were formatted using a system-message (emphasizing the required behavior) followed by a user-message structure.

While the specific content of the prompts varies depending on the role and task (e.g., simulating user intents, extracting slots, managing dialogue state transitions), the overall prompt structure remains consistent across all scenarios. Each prompt is composed of the following components:
\begin{enumerate}
    \item System Message: Defines the role that the LLM is expected to play (e.g., user simulator, dialogue manager, slot extractor).

    \item Task Description: Provides a brief overview of the task context and the objective the LLM should achieve.

    \item Instructions and Rules: Specifies detailed guidelines and constraints that the LLM must follow when generating responses.

    \item Test Input: The specific input query for which a response is to be generated.
\end{enumerate}

This prompt format ensures that models are contextualized for their assigned tasks while allowing flexibility in content to accommodate the differing requirements of user simulation, dialogue management, and modular dialogue tasks.

\begin{table*}[t]
  \begin{minipage}{\textwidth}
     \centering
    \scriptsize
    \begin{tabular}{lcccccccc}
      \toprule
        \multirow{2}{*}{\textbf{Model}} & \multicolumn{4}{c}{\textbf{Domain Level}} & \multicolumn{4}{c}{\textbf{Dialogue Level}} \\
        & \textbf{Inform} & \textbf{Success} & \textbf{Book} & \textbf{Combine} &\textbf{Inform} & \textbf{Success} & \textbf{Book} & \textbf{Combine} \\
        \midrule
          \verb|Llama-2-13B|   & 0.37 & 0.29 & 0.32 & 0.34 & 0.29 & 0.23 & 0.27 & 0.27 \\
          \verb|Llama-2-70B|   & 0.54 & 0.43 & 0.44 & 0.49 & 0.33 & 0.31 & 0.32 & 0.32 \\
          \verb|GPT-3.5-Turbo|  & 0.63 & 0.53 & 0.51 & 0.57 & 0.43 & 0.46 & 0.48 & 0.46 \\
          \verb|GPT-4o|  & \textbf{0.85} & \textbf{0.59} & \textbf{0.87} & \textbf{0.79} & \textbf{0.80} & \textbf{0.47} & \textbf{0.82} & \textbf{0.72} \\
      \bottomrule
    \end{tabular}
    \captionof{table}{Domain-level and dialogue-level performance metrics (Inform, Success, Book, and Combine) that were reported in the baseline paper~\citep{DBLP:conf/acl/XuMYSH24}} 
    \label{tab:xe_baseline_paper}
  \end{minipage}
   \vfill 
  \begin{minipage}{\textwidth}
     \centering
      \scriptsize    
    \begin{tabular}{lcccccc}
        \toprule
        \multirow{2}{*}{\textbf{Model}} & \multicolumn{3}{c}{\textbf{Domain Level}} & \multicolumn{3}{c}{\textbf{Dialogue Level}} \\
        & \textbf{Inform} & \textbf{Book} & \textbf{Combine} &\textbf{Inform} & \textbf{Book} & \textbf{Combine} \\
        \midrule
          \verb|Llama-2-13B|   & 0.00 & 0.00 & 0.00 & 0.00 & 0.00 & 0.00 \\
          \verb|Llama-2-70B|   & 0.00 & 0.00 & 0.00 & 0.00& 0.00 & 0.00 \\
          \verb|Llama-3.2-1B|   & 0.36 & 0.00 & 0.13 & 0.27 & 0.00 & 0.08 \\
          \verb|Llama-3.2-3B|   & 0.24 & 0.00 & 0.09 & 0.09 & 0.00 & 0.03 \\
          \verb|Qwen2.5-7B|     & 0.11 & 0.00 & 0.05 & 0.04 & 0.00 & 0.02 \\
          \verb|Llama-3.1-8B|   & 0.17 & 0.00 & 0.07 & 0.1 & 0.00 & 0.05 \\
          \verb|Qwen2.5-32B|    & 0.11 & 0.00 & 0.05 & 0.05 & 0.00 & 0.03 \\
          \verb|Llama-3.3-70B|  & 0.11 &  0.00 & 0.06 & 0.06 & 0.00 & 0.03 \\
          \verb|GPT-4o|  & \textbf{0.73} & \textbf{0.73} & \textbf{0.59} & \textbf{0.67} & \textbf{0.67} & \textbf{0.51} \\
        \bottomrule
      \end{tabular}
    \captionof{table}{Results of the baseline system on the filtered test set. We use the same model for both the user and agent. As the results are based on a filtered subset, performance differs from the baseline but comparable for \texttt{GPT-4o}.} 
    \label{tab:xe_setup_same_models}
  \end{minipage}
   \vfill 
  \begin{minipage}{\textwidth}
     \centering
      \scriptsize    
    \begin{tabular}{lcccccc}
        \toprule
        \multirow{2}{*}{\textbf{Model}} & \multicolumn{3}{c}{\textbf{Domain Level}} & \multicolumn{3}{c}{\textbf{Dialogue Level}} \\
        & \textbf{Inform} & \textbf{Book} & \textbf{Combine} &\textbf{Inform} & \textbf{Book} & \textbf{Combine} \\
        \midrule
          \verb|Llama-2-13B|   & 0.00 & 0.00 & 0.00 & 0.00 & 0.00 & 0.00 \\
          \verb|Llama-2-70B|   & 0.00 & 0.00 & 0.00 & 0.00 & 0.00 & 0.00 \\
          \verb|Llama-3.2-1B|   & 0.40 & 0.00 & 0.13 & 0.23 & 0.00 & 0.07 \\
          \verb|Llama-3.2-3B|   & 0.24 & 0.00 & 0.09 & 0.09 & 0.00 & 0.03 \\
          \verb|Qwen2.5-7B|     & 0.11 & 0.00 & 0.05 & 0.04 & 0.00 & 0.02 \\
          \verb|Llama-3.1-8B|   & 0.18 & 0.00 & 0.07 & 0.11 & 0.00 & 0.04 \\
          \verb|Qwen2.5-32B|    & 0.11 & 0.00 & 0.05 & 0.05 & 0.00 & 0.03 \\
          \verb|Llama-3.3-70B|  & 0.12 &0.00 & 0.06 & 0.06 & 0.00 & 0.03 \\
          \verb|GPT-4o|  &  0.16 & 0.00 & 0.07 & 0.10 & 0.00 & 0.04 \\
        \bottomrule
      \end{tabular}
    \captionof{table}{Baseline system results using \texttt{Qwen-2.5-32B} as the user simulator and varying the agent model. The performance drops significantly compared to using the same model for both user and agent, showcasing the sensitivity of the dialogue system to the user simulator.} 
    \label{tab:xe_setup_same_usmodel}
  \end{minipage}   
\end{table*}

\subsection {Existing Systems Integration}
\label{sec:appendix-existing-sys-integration}
As part of our benchmarking, we integrated two representative task-oriented dialogue systems from prior work into the \clemtodd\  framework: AutoTOD~\citep{DBLP:conf/acl/XuMYSH24} and the modular pipeline proposed by ~\citet{DBLP:conf/sigdial/HudecekD23}.

\subsubsection{AutoTOD Integration}
AutoTOD is a zero-shot, monolithic system, using a LangChain-powered agent. In our integration, we maintained AutoTOD’s internal logic for database operations and booking reference number generation to maintain consistency with its original design. However, we ensured that it operated over the same MultiWOZ 2.2 database used for all other dialogue systems in our framework, thus preserving fairness in task execution and evaluation.

AutoTOD's original implementation~\footnote{\url{https://github.com/DaDaMrX/AutoTOD}} uses an older version of LangChain (v$0.0.166$), which required adjustments to ensure compatibility with current LangChain APIs. The framework includes two agent types: a ReAct agent and a function agent. We initially attempted to use the ReAct agent, but integration failed due to format mismatches between the LLM-generated outputs and the expected tool call structure—an issue stemming from changes introduced in newer LangChain versions. As a result, we integrated the function agent, which was working under the upgraded environment.

\begin{table}[t]
  \begin{minipage}{0.48\textwidth}
     \centering
    \scriptsize
    \begin{tabular}{lccc}
      \toprule
        \multirow{2}{*}{\textbf{Model}} & \textbf{Inform} & \textbf{Success} & \textbf{BLEU}\\
        \midrule
          \verb|ChatGPT-Zeroshot|   & 0.47 & NA & 3.76 \\
          \verb|ChatGPT-Fewshot|   & 0.68 & NA & 6.84\\
      \bottomrule
    \end{tabular}
    \captionof{table}{Performance metrics (Inform, Success, and Book) that were reported in the baseline paper~\citep{DBLP:conf/sigdial/HudecekD23}} 
    \label{tab:he_baseline_paper}
  \end{minipage}  
\end{table}

The function agent, however, expects LLMs to generate raw SQL queries. This introduced some issues, as the case of generated column names occasionally did not match the schema of the underlying database. To address this, we modified the database layer to support case-insensitive column matching. Additionally, the booking function uses regular expressions to extract timing information in train booking scenarios. These expressions were failing in some edge cases and we therefore updated the relevant regex patterns.

In Table~\ref{tab:xe_baseline_paper}, we report the AutoTOD results from the original paper. To enable fair comparison, we reproduced the system in its original setup and evaluated in two modes inspired by our \clemtodd\ experiments: first, by using the same model for both the agent and the dialogue system, and second, by using a single instruction-following model \texttt{Qwen2.5-32B} as user model against other models for dialogue systems. These results are presented in Table~\ref{tab:xe_setup_same_usmodel}.

Our analysis indicates that performance improves when the same model is used across components, as opposed to mixing models. Furthermore, when using \texttt{Qwen2.5-32B} based user simulator in our setup, the dialogue system outperformed its own setup, likely due to the improved user simulator prompts provided by \clemtodd\. We also observe a performance gap between the paper-reported results and our reproduced results, which is likely due to differences in agent configuration during evaluation, the original paper might have used the ReAct agent, whereas we used the function agent due to compatibility issues with newer LangChain versions. This change in agent behavior may have affected how tool calls were handled and how robustly dialogue goals were completed.

\begin{table}[ht]
    \begin{minipage}{0.48\textwidth}
     \centering
      \scriptsize    
    \begin{tabular}{lccccc}
        \toprule
        {\textbf{Model}} & \textbf{BLEU} & \textbf{Inform} & \textbf{R} & \textbf{H} & \textbf{T}\\
        \midrule
          \verb|Llama-2-13B|   & 2.43 & 0.30 & 0.18 & 1.00 & 0.25 \\
          \verb|Llama-2-70B|   & 0.58 & 0.30 & 0.18 & 1.00 & 0.25 \\
          \verb|Llama-3.2-1B|   & 0.32 & 0.30 & 0.18 & 1.00 & 0.25 \\
          \verb|Llama-3.2-3B|   & 0.33 & 0.30 & 0.18 & 1.00 & 0.25 \\
          \verb|Qwen2.5-7B|     & 0.54 & 0.30 & 0.18 & 1.00 & 0.25 \\
          \verb|Llama-3.1-8B|   & 0.49 & 0.30 & 0.18 & 1.00 & 0.25 \\
          \verb|Qwen2.5-32B|    & 0.65 & 0.30 & 0.18 & 1.00 & 0.25 \\
          \verb|Llama-3.3-70B|  & 0.58 & 0.30 & 0.18 & 1.00 & 0.25 \\
          \verb|GPT-4o|  &  0.73 & 0.30 & 0.18 & 1.00 & 0.25  \\
        \bottomrule
      \end{tabular}
    \captionof{table}{Few-shot results on the Multiwoz filtered test set using the same model for both user and dialogue system. Metrics include overall Inform, and domain-specific scores for restaurant (R), hotel (H), and train (T). Performance differs from the baseline due to filtering.} 
    \label{tab:xe_setup_same_usmodel}
  \end{minipage}  
\end{table}

\subsubsection{Modular Pipeline System Integration}
We integrated the modular pipeline dialogue system proposed by~\citet{DBLP:conf/sigdial/HudecekD23}. The system comes with its own implementation~\footnote{\url{https://github.com/vojtsek/to-llm-bot}} for handling database interactions and generating booking reference numbers. We retained these components as is but ensured that the underlying database matched the MultiWOZ 2.2 version used across all systems in \clemtodd\ to maintain consistency.

The original implementation used delexicalized responses, as their evaluation setup did not rely on realistic user simulation. Since \clemtodd\ requires naturalistic dialogue for interaction with LLM-based user simulators, we modified the code to disable delexicalization, allowing entity names and values to appear directly in the system responses. Additionally, their regular expression logic for extracting domains from user input did not generalize well across all test cases, so we extended this pattern to improve robustness.

The original system also used the Weights \& Biases (wandb) framework for logging and tracking experiments. As this was not essential for integration with \clemtodd\, we removed these dependencies during setup. Furthermore, the original implementation did not compute booking success as part of its evaluation, so we could not report these results.

\begin{figure*}
    \includegraphics[scale=0.74]{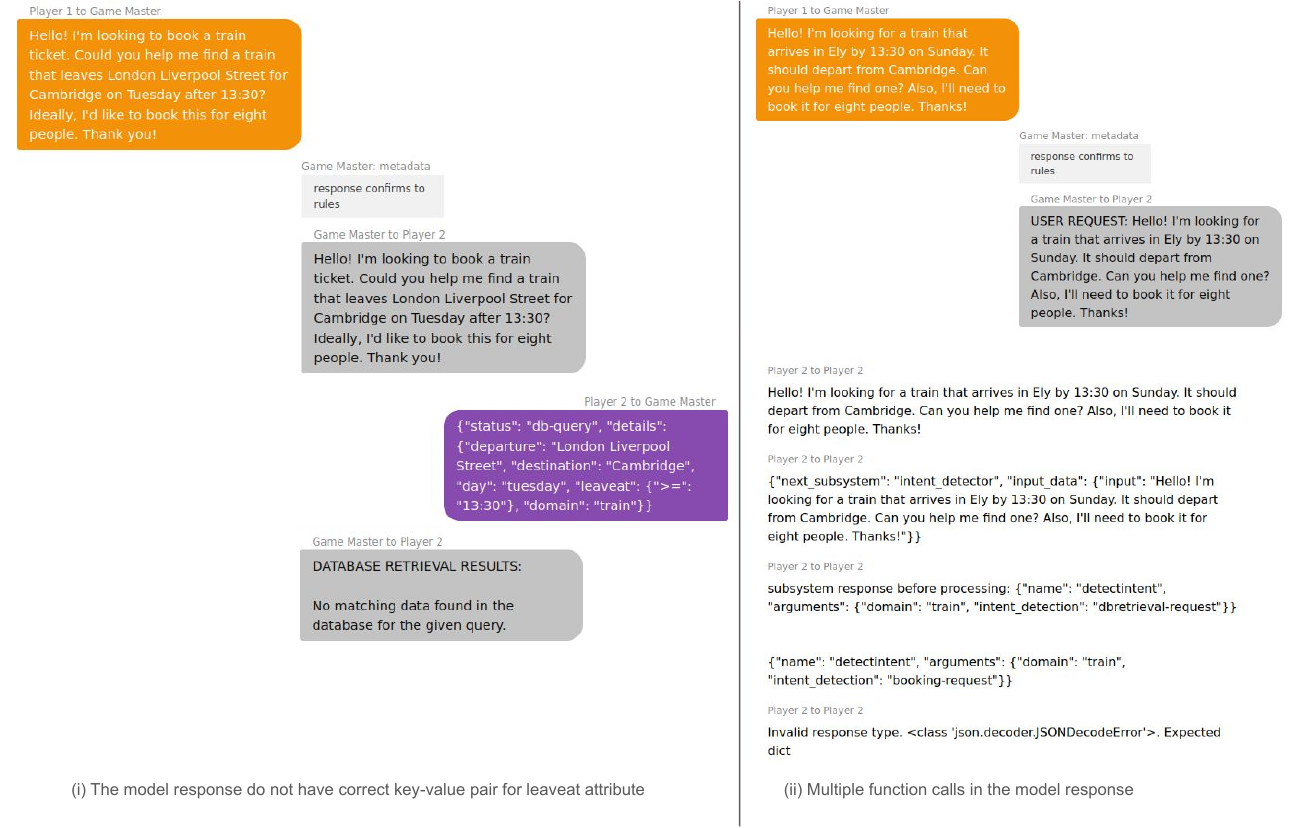}
    \captionof{figure}{Examples of invalid model responses due to format violations during dialogue system evaluation.}
    \label{fig:ds_invalid_response}
\end{figure*}
\label{sec:response-format-issues}

\subsection {Response Format Issues}
Despite explicitly defining constraints and tool schemas for response generation, models do not always adhere strictly to the specified formats. Deviations from the expected response format often lead to invalid outputs, resulting in parsing errors or task execution failures, which ultimately reduce the overall evaluation scores.

common types of invalid responses include incorrect key-value structures, missing required fields, and multiple function calls when only a single function call was expected. Figure~\ref{fig:ds_invalid_response} illustrates such invalid responses observed during evaluation.

\subsection {Dialogue Quality Evaluation}
\label{sec:appendix-dlgeval}
To assess the quality of the generated user simulator utterances, we follow the methodology outlined by \citet{DBLP:conf/slt/KaziLZHT24}. Specifically, we use two models in a zero-shot setting to score the utterances: a closed-source model (\verb|GPT-4o|) and an open-weight model (\verb|LLaMA-3.3-70B|). The evaluation focuses on three dimensions: naturalness (N), coherence (C), and dialogue-level diversity (D). Naturalness is rated on a scale from 1 to 5, while coherence and diversity are rated on a scale from 1 to 3, with higher scores indicating better performance.

To cross-validate these \verb|GPT-4o| findings(see Table~\ref{tab:usdlgquality}), we conducted a parallel evaluation using \verb|LLaMA-3.3-70B|, and results are shown in Table~\ref{tab:usdlgquality_l370b}. The open-weight model exhibits similar trends, reinforcing the observations made with \verb|GPT-4o|.

\begin{table}[ht]
  \begin{minipage}[t]{0.48\textwidth}
      \vspace{0pt} 
      \scriptsize
        \begin{tabular}{lccc}
          \toprule
          {\textbf{Model}} & Naturalness & Coherence & Dialogue Diversity \\
          \midrule
          \verb|Llama-3.2-1B|   & 3.90 & 2.90 & 1.00 \\
          \verb|Llama-3.2-3B|   & 3.25 & 2.18 & 1.02 \\
          \verb|Qwen2.5-7B|    & 2.55 & 2.22 & 1.12 \\
          \verb|Llama-3.1-8B|   & 2.97 & 2.32 & 1.03 \\
          \verb|Qwen2.5-32B|   & \textbf{4.87} & \textbf{3.00} & 1.00 \\
          \verb|Llama-3.3-70B|  & 4.28 & 2.93 & 1.00 \\
          \bottomrule          
        \end{tabular}
        \caption{Dialogue quality comparison of user simulators for the Monolithic architecture-based dialogue system (using the model: \texttt{Qwen2.5-32B}), evaluated on Naturalness (N), Coherence (C), and Dialogue Diversity (D) metrics using \texttt{Llama-3.3-70B}. Higher scores indicate better performance.}
        \label{tab:usdlgquality_l370b}
  \end{minipage}
\end{table}

\begin{figure*}
    \includegraphics[scale=0.65]{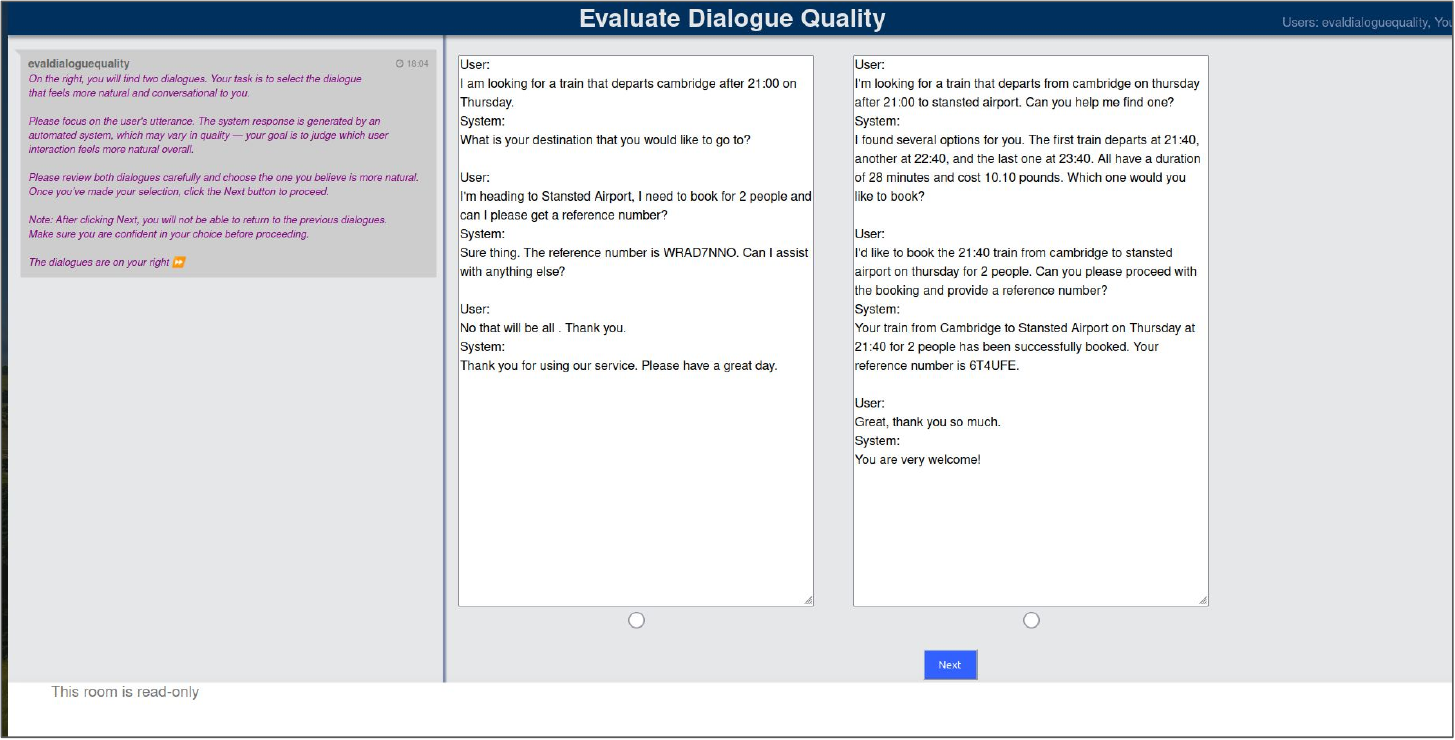}
    \captionof{figure}{Web interface used for the Human Turing Test. Annotators were shown two dialogues side by side (a generated dialogue and a ground-truth dialogue) and asked to select the one that appeared more natural.
    }
    \label{fig:usturingtest_interface}
\end{figure*}

Dialogue diversity scores remain mostly stable, typically around $\sim 1.0$, across different models and architectures. This suggests limited variability in simulator utterances. While task-oriented dialogues are inherently more constrained than open-domain dialogues, real users often introduce a degree of natural variation~\citep{DBLP:journals/corr/abs-2305-13857} through paraphrasing, hesitations, and reformulations. The relatively low diversity observed here implies that current user simulators may be overly deterministic, potentially restricting the robustness of dialogue systems trained with them. To address this, future work should aim to develop user simulators that balance fluency with controlled diversity, thereby better approximating real-world conversational behaviors.

\paragraph {Human Evaluation}
\label{sec:appendix-dlgeval-humaneval}
To further investigate the performance of the user simulator, we conducted a Turing Test to compare generated dialogues against ground-truth dialogues. 
\begin{table*}
\centering
\scriptsize
    \begin{tabular}{clcccccccccccc}
      \hline
      \multirow{2}{*}{Domain} & \multirow{2}{*}{Model} & \multicolumn{4}{c}{\textbf{Monolithic}}& \multicolumn{4}{c}{\textbf{Modular-Prog}} & \multicolumn{4}{c}{\textbf{Modular-LLM}} \\
                     & & R & H & T & Avg & R & H & T & Avg & R & H & T & Avg\\
      \hline
      \multirow{7}{*}{Single} & \verb|Llama-3.2-1B| & 0.00 & 0.00 & 0.00 & 0.00 & 0.00 & 0.00 & 0.00 & 0.00 & 0.00 & 0.00 & 0.00 & 0.00 \\
      & \verb|Llama-3.2-3B| & 0.20 & 0.05 & 0.00 & 0.08 & 0.00 & 0.00 & 0.00 & 0.00 & 0.10 & 0.00 & 0.00 & 0.03 \\
      & \verb|Qwen2.5-7B| & 0.00 & 0.00 & 0.30 & 0.10 & 0.70 & 0.50 & 0.20 & 0.47 & 0.45 & 0.25 & 0.20 & 0.30 \\
      & \verb|Llama-3.1-8B| & 0.50 & 0.25 & 0.10 & 0.28 & 0.00 & 0.00 & 0.00 & 0.00 & 0.20 & 0.10 & 0.00 & 0.10 \\
      & \verb|Qwen2.5-32B| & 1.00 & 0.90 & 0.95 & \textbf{0.95} & 0.80 & 0.50 & 0.45 & 0.58 & 0.95 & 0.75 & 0.75 & 0.82 \\
      & \verb|Llama-3.3-70B| & 0.85 & 0.65 & 1.00 & 0.83 & 0.80 & 0.35 & 0.45 & 0.53 & 0.80 & 0.50 & 0.80  & 0.70\\
      & \verb|GPT-4o| & 0.95 & 0.70 & 0.50 & 0.72 & 1.00 & 0.80 & 0.75 & \textbf{0.85} & 1.00 & 0.95 & 0.65 & \textbf{0.87} \\
      \hline
      Domain & Model & R-H & H-T & T-R & Avg & R-H & H-T & T-R & Avg & R-H & H-T & T-R & Avg \\
      \hline                     
      \multirow{7}{*}{Multi} & \verb|Llama-3.2-1B| & 0.00 & 0.00 & 0.00 & 0.00 & 0.80 & 0.50 & 0.45 & 0.58 & 0.00 & 0.00 & 0.00 & 0.00 \\
      & \verb|Llama-3.2-3B| & 0.00 & 0.00 & 0.05 & 0.02 & 0.80 & 0.50 & 0.45 & 0.58 & 0.00 & 0.00 & 0.00 & 0.00 \\
      & \verb|Qwen2.5-7B| & 0.00 & 0.10 & 0.10 & 0.07 & 0.80 & 0.50 & 0.45 & 0.58 & 0.19 & 0.05 & 0.25 & 0.16 \\
      & \verb|Llama-3.1-8B| & 0.18 & 0.00 & 0.15 & 0.11 & 0.80 & 0.50 & 0.45 & 0.58 & 0.05 & 0.00 & 0.05 & 0.02 \\
      & \verb|Qwen2.5-32B| & 0.71 & 0.45 & 0.60 & 0.59 & 0.24 & 0.15 & 0.25 & 0.25 & 0.59 & 0.50 & 0.55 & 0.55 \\
      & \verb|Llama-3.3-70B| & 0.47 & 0.40 & 0.60 & 0.49 & 0.41 & 0.80 & 0.35 & 0.45 & 0.47 & 0.25 & 0.30 & 0.34 \\
      & \verb|GPT-4o| & 0.76 & 0.45 & 0.90 & \textbf{0.70} & 1.00 & 0.80 & 0.75 & \textbf{0.56} & 0.94 & 0.70 & 0.80 & \textbf{0.81} \\
      \hline
    \end{tabular}
  \caption{Booking rates for each model across different dialogue system architectures (Monolithic, Modular-Prog, Modular-LLM) and domains. ``Single'' refers to tasks within a single domain (Restaurant(R), Hotel(H), or Train(T)), while ``Multi'' refers to tasks spanning multiple domains. Higher values indicate better task completion performance.}
  \label{tab:domainwise_results}
\end{table*}

We recruited an annotator to perform the annotation. The annotator did not have prior experience in dialogue annotation but had general experience with data labeling tasks. Detailed instructions for the Turing Test were embedded within the web interface (see Figure~\ref{fig:usturingtest_interface}) to guide the evaluation process. The annotation task involved evaluating 100 dialogue pairs and took approximately 150 minutes to complete.

The results are summarized in Table~\ref{tab:usdlgquality} under TT (Turing Test) metric. For each model, 50 generated dialogues were evaluated against ground-truth dialogues. Out of 50 comparisons, 19 dialogues generated by \verb|Qwen2.5-32B| were judged to be more natural, while only 8 dialogues generated by \verb|LLaMA-3.3-70B| were preferred.

It is important to note that the human Turing Test and the LLM-judge evaluation use fundamentally different methodologies. The human evaluation compares generated dialogues directly against ground-truth dialogues for the same task, whereas the LLM-judge evaluation scores each generated dialogue independently based on naturalness, coherence, and diversity, without referencing the ground truth. Although the results from the two evaluations are not directly comparable, both indicate that the dialogues generated by the \verb|Qwen2.5-32B| model tend to sound more natural.

\subsection {Domain-specific Results}
\label{sec:appendix-domain-results}
These domain-wise results provide a finer-grained view of how model and system design choices impact task completion across different dialogue scenarios. Table~\ref{tab:domainwise_results} presents detailed booking success rates for each model across different dialogue system architectures (Monolithic, Modular-Prog, and Modular-LLM) and task settings (Single-domain and Multi-domain). For single-domain tasks, the performance is reported separately for Restaurant (R), Hotel (H), and Train (T) domains, along with their average. For multi-domain tasks, performance is reported for each domain pair (e.g., Restaurant–Hotel (R-H), Hotel–Train (H-T), Train–Restaurant (T-R)), along with the corresponding average.

Across architectures, models generally achieve higher booking rates on single-domain tasks compared to multi-domain tasks, reflecting the increased complexity introduced by multi-domain interactions. For single-domain tasks, the majority of failures are due to the model not adhering strictly to the specified response format constraints. For multi-domain tasks, in addition to format violations, several other issues contribute to performance degradation. These include user simulator occasionally terminating the conversation prematurely after completing only one task, and models assuming missing information (such as hotel area or restaurant food type) that differed from the ground-truth data, resulting in different booking choices and thus reduced the overall scores.

\subsection {Cost Estimation}
\label{sec:appendix-cost-estimation}
We estimate computational costs across dialogue system architectures and model scales by analyzing both token-based and FLOP-based expenses. This evaluation provides insights into the computational trade-offs associated with different dialogue system architectures.

We followed a standardized estimation methodology~\citep{DBLP:journals/corr/abs-2001-08361, DBLP:journals/jmlr/ChowdheryNDBMRBCSGSSTMRBTSPRDHPBAI23} to compute the number of tokens, floating-point operations (FLOPs), and their associated costs.

\paragraph{Token Cost Computation} Although all open-weight models were run locally, token costs were estimated using pricing information from OpenRouter APIs~\footnote{\url{https://openrouter.ai/models}} to provide an indicative view of real-world deployment expenses. The total token cost (T) is computed as the sum of the prompt(p) and response(r) token costs, using the respective input ($c_{i}$) and output ($c_{o}$) pricing rates for each model:

\begin{equation*}
T = (p \times c_i) + (r \times c_o)
\end{equation*}

\begin{table*}
  \begin{minipage}[t]{0.48\textwidth}
    \vspace{0pt} 
    \centering
      \scriptsize
    \begin{tabular}{lccccc}
      \hline
      Domain & Model & Restaurant & Hotel & Train & Avg \\
      \hline
      \multirow{7}{*}{Single} & \verb|Llama-3.2-1B| & 0.00 & 0.00 & 0.00 & 0.00 \\
      & \verb|Llama-3.2-3B| & 0.35 & 0.00 & 0.65 & 0.33 \\
      & \verb|Qwen2.5-7B| & 0.20 & 0.10 & 0.40 & 0.23 \\
      & \verb|Llama-3.1-8B| & 0.82 & 0.25 & 0.65 & 0.57\\
      & \verb|Qwen2.5-32B| & 0.90 & 0.80 & 0.90 & \textbf{0.87} \\
      & \verb|Llama-3.3-70B| & 0.70 & 0.50 & 0.85 & 0.68 \\
      & \verb|GPT-4o| & 0.80 & 0.60 & 0.90 & 0.77 \\
      \hline
      Domain & Model & R-H & H-T & T-R & Avg \\
      \hline                     
      \multirow{7}{*}{Multi} & \verb|Llama-3.2-1B| & 0.00 & 0.00 & 0.00 & 0.00 \\
      & \verb|Llama-3.2-3B| & 0.00 & 0.00 & 0.05 & 0.02 \\
      & \verb|Qwen2.5-7B| & 0.00 & 0.00 & 0.10 & 0.03 \\
      & \verb|Llama-3.1-8B| & 0.00 & 0.00 & 0.00 & 0.00 \\
      & \verb|Qwen2.5-32B| & 0.35 & 0.35 & 0.35 & 0.42 \\
      & \verb|Llama-3.3-70B| & 0.30 & 0.20 & 0.70 & 0.40 \\
      & \verb|GPT-4o| & 0.35 & 0.45 & 0.75 & \textbf{0.52} \\
      \hline
    \end{tabular}
  \caption{Performance of monolithic dialogue systems evaluated with Qwen2.5-32B as the user simulator on the MultiWOZ-style synthetic dataset. Results are reported for single-domain (Restaurant, Hotel, Train) and multi-domain (Restaurant-Hotel, Hotel-Train, Train-Restaurant) tasks. ``Avg'' denotes the average score across the domains.}
  \label{tab:synthetic-multiwozstyle-domainwise_results}
  \end{minipage}
  \hfill
  \begin{minipage}[t]{0.48\textwidth}
    \vspace{0pt} 
    \centering
      \scriptsize
    \begin{tabular}{lccccc}
      \hline
      Domain & Model & Restaurant & Hotel & Train & Avg \\
      \hline
      \multirow{7}{*}{Single} & \verb|Llama-3.2-1B| & 0.00 & 0.00 & 0.00 & 0.00 \\
      & \verb|Llama-3.2-3B| & 0.00 & 0.00 & 0.10 & 0.03 \\
      & \verb|Qwen2.5-7B| & 0.05 & 0.05 & 0.05 & 0.05 \\
      & \verb|Llama-3.1-8B| & 0.20 & 0.05 & 0.15 & 0.13\\
      & \verb|Qwen2.5-32B| & 0.10 & 0.50 & 0.25 & 0.28 \\
      & \verb|Llama-3.3-70B| & 0.05 & 0.05 & 0.20 & 0.10 \\
      & \verb|GPT-4o| & 0.15 & 0.80 & 0.40 & \textbf{0.45} \\
      \hline
      Domain & Model & R-H & H-T & T-R & Avg \\
      \hline                     
      \multirow{7}{*}{Multi} & \verb|Llama-3.2-1B| & 0.00 & 0.00 & 0.00 & 0.00 \\
      & \verb|Llama-3.2-3B| & 0.00 & 0.00 & 0.00 & 0.00 \\
      & \verb|Qwen2.5-7B| & 0.00 & 0.00 & 0.00 & 0.00 \\
      & \verb|Llama-3.1-8B| & 0.00 & 0.00 & 0.00 & 0.00 \\
      & \verb|Qwen2.5-32B| & 0.10 & 0.10 & 0.00 & 0.07 \\
      & \verb|Llama-3.3-70B| & 0.00 & 0.05 & 0.05 & 0.03 \\
      & \verb|GPT-4o| & 0.15 & 0.45 & 0.05 & \textbf{0.22} \\
      \hline
    \end{tabular}
  \caption{Performance of monolithic dialogue systems evaluated with Qwen2.5-32B as the user simulator on the unrealistic synthetic dataset. Results are reported for single-domain (Restaurant, Hotel, Train) and multi-domain (Restaurant-Hotel, Hotel-Train, Train-Restaurant) tasks. ``Avg'' denotes the average score across the domains.}
  \label{tab:synthetic-unrealistic-domainwise_results}
  \end{minipage}      
\end{table*}

\paragraph{Flop Cost Computation} The FLOP cost (F) is derived by first estimating the total number of floating-point operations (FP) required to process all tokens, using an estimate~\citep{DBLP:journals/corr/abs-2001-08361, DBLP:journals/jmlr/ChowdheryNDBMRBCSGSSTMRBTSPRDHPBAI23} of $2 \times$ the number of model parameters per token. This value is then converted into petaflops and priced based on a standard cost assumption. Specifically, we assume a cost of \$0.05 per petaFLOP ($c_{pf}$), derived from an A100 GPU rental rate of approximately \$2 per hour. This allows us to estimate the computational overhead of inference independently of specific deployment hardware.

\begin{align*}
FP = 2 \times \text{model parameters}\\
\text{Total PetaFLOPs} =  \frac{(p \times FP) + (r \times FP)}{10^{15}} \\
F = \text{Total PetaFLOPs} \times c_{pf}
\end{align*}

\subsection{Synthetic Dataset Generation}
\label{sec:appendix-synthetic-data-generation}
To assess the adaptability and generalization capabilities of the \clemtodd\ framework, we created two types of synthetic test datasets: one that mimics the structure of the original MultiWOZ 2.2 benchmark but introduces unseen goal configurations, and another that contains intentionally unrealistic or adversarial dialogue tasks.
\subsubsection{MultiWOZ-Style Synthetic Dataset}
This dataset preserves the general structure and domain constraints of MultiWOZ 2.2 (restaurant, hotel, and train), but combines slot values and goals in configurations not found in the original corpus. The goal was to simulate unseen but similar tasks to assess how well systems generalize.

To construct the MultiWOZ-style synthetic dataset, we first collected the available slots for each domain (restaurant, hotel, and train) and then randomly generated novel combinations of these slots. Each generated goal was checked to ensure that the exact same combination of slot types and values did not exist in the original MultiWOZ 2.2 corpus, thereby guaranteeing the inclusion of unseen goal configurations. These synthetic goals were then rendered into natural language using templates. We generated a total of 120 task goals to match the size of the baseline evaluation set, comprising 60 single-domain and 60 multi-domain dialogue tasks.

\subsubsection{Unrealistic Synthetic Dataset}
For the unrealistic dataset, we constructed a custom ontology that retained the original slot names from the MultiWOZ dataset, such as area, food, hotel type, and travel day etc., but replaced their values with unusual values. For example, area values included ``middle of ocean'' and ``top of volcano'', hotel type included ``dungeon'' and ``wormhole'', food included ``rotten'' or ``leftover'', and travel days were set as ``someday'' or ``yesterday''. Using this ontology, we generated random combinations of slot-value pairs and then rendered them into natural language goals using the same template-based generation approach as the synthetic MultiWOZ-style dataset. We created a total of 120 tasks, 60 single-domain and 60 multi-domain, to mirror the scale of the baseline evaluation set.  These tasks, while structurally valid, were semantically unrealistic, allowing us to probe the dialogue systems’ behavior in the face of adversarial user requests.

\begin{figure*}
  \centering
    \begin{prompt}
\\
\textbf{System Info}
\\\\
ROLE: You are a user simulator tasked with interacting naturally with a dialogue system.
\\\\
\textbf{TASK:}
\\\\
\$goal
\\\\
\textbf{INSTRUCTIONS:}
\\\\
1. Communicate naturally by expressing preferences, asking clarifying questions, and making decisions as needed.

2. Maintain a polite and conversational tone.

3. Respond strictly based on the dialogue system's response. Do not add logic or interpretation beyond what is explicitly stated in the TASK.

4. Ensure that names and terms remain exactly as provided in the input, without any added or altered punctuation (e.g., do not add apostrophes, hyphens, or other symbols). Maintain strict adherence to the original formatting.

5. When booking a train, exact time matches may not always be available. If the dialogue system provides alternative options close to the desired time, the you should accept a suitable nearby option that reasonably aligns with the goal.

6. Once the dialogue system completes the task and provides the reference number, reply with "DONE". No additional text should follow/preceed.

7. Do not simulate or act as the dialogue system; only interact with it.

8. Keep responses concise and focused, avoiding unnecessary elaboration or overly conversational tone.
\\\\
\textbf{OUTPUT FORMAT}
\\\\
1. Interaction: Respond appropriately using only the dialogue system's response and the information under TASK.

2. Task Completion: Reply with "DONE". Do not add any customary comments (thank you, great etc.)

3. Use 'DONE' only (without any prefix/suffix) to confirm task completion.

Lets begin
\end{prompt}
\caption{Prompt template for the User Simulator, specifying the task description, interaction instructions, and response format guidelines.}
    \label{fig:usimulator_prompt}
\end{figure*}
\begin{figure*}
  \centering
    \begin{prompt}
\textbf{System Info}
\\\\
ROLE: You are a specialized booking assistant interacting with a human user through JSON function calls using the provided tool schema. Your role is to process user requests and ensure successful task completion while maintaining a professional, helpful tone.  You are NOT allowed to return free-form messages outside tool calls.
\\\\
\textbf{TASK:}
\\\\
Assist the user conversationally by:

1. Extracting key details needed for the task (e.g., domain, date, time, location).

2. Cross-referencing user-provided information with the database to find relevant matches.

3. If too many records are available, the database system returns only the first five. If the required information is not available in the returned records, apply additional filters to narrow down the results.

4. Generating responses to gather missing or unclear information or to provide the booking status.

5. For train bookings, if the database does not have trains available at the exact requested time, clarify with the user whether they are interested in seeing the closest available options that best match their query.

6. Consolidating all extracted and clarified details for booking finalization.

7. Keeping responses concise and focused, avoiding unnecessary elaboration or overly conversational tone.

8. Do not assume any details; always ask the user for clarification when necessary.
\\\\
\textbf{INSTRUCTIONS:}
\\\\
1. Communicate naturally by expressing preferences, asking clarifying questions, and making decisions as needed.

2. Maintain a polite and conversational tone.

3. Respond strictly based on the dialogue system's response. Do not add logic or interpretation beyond what is explicitly stated in the TASK.

4. Ensure that names and terms remain exactly as provided in the input, without any added or altered punctuation (e.g., do not add apostrophes, hyphens, or other symbols). Maintain strict adherence to the original formatting.

5. When booking a train, exact time matches may not always be available. If the dialogue system provides alternative options close to the desired time, the you should accept a suitable nearby option that reasonably aligns with the goal.

6. Once the dialogue system completes the task and provides the reference number, reply with "DONE". No additional text should follow/preceed.

7. Do not simulate or act as the dialogue system; only interact with it.

8. Keep responses concise and focused, avoiding unnecessary elaboration or overly conversational tone.
\\\\
\textbf{OUTPUT FORMAT}
\\\\
1. Interaction: Respond appropriately using only the dialogue system's response and the information under TASK.

2. Task Completion: Reply with "DONE". Do not add any customary comments (thank you, great etc.)

3. Use 'DONE' only (without any prefix/suffix) to confirm task completion.

Lets begin
\\\\
\$USER\_SIMULATOR\_UTTERANCE
\end{prompt}
\caption{Prompt template for the monolithic dialogue system, detailing the task procedures, interaction instructions, and output format for user request processing via JSON function calls.}
    \label{fig:monods_prompt}
\end{figure*}
\begin{figure*}
  \centering
    \begin{prompt}
\textbf{System Info}
\\\\
ROLE: You are the dialogue manager for a specialized booking assistant bot and interact through JSON function calls using the provided tool schema. Your role is to process user requests, coordinate interactions with subsystems, and ensure successful task completion. You are NOT allowed to return free-form messages outside tool calls.
\\\\
\textbf{TASK:}
\\\\
1. For each user request:

   a. Determine appropriate flow based on user input and available information

   b. Identify next required subsystem. Always use the exact subsystem names as specified in the tool 
   schema.

   c. Prepare the necessary input data for that subsystem

   d. For database queries and validating booking information, use the exact function names as specified in the tool schema.

   e. Do not generate any booking confirmation (reference number) on your own. Use the appropriate function to validate the booking and to generate the reference number.

2. All responses must strictly adhere to the format. Include all required fields and the response must be a valid JSON.
\\\\
\textbf{RESPONSE RULESS:}
\\\\
1. Use the most appropriate function call based on the user’s request and available data.

2. To interact between the sub-systems (intent detection, slot extraction, or response generation), call the `processnextsubsystem` function.

3. To respond to the user, as a final message after coordinating with the dialogue sub-systems call the `followup` function.

4. Similarly for booking action or database lookup, use the appropriate function from the tool schema.

5. Every response MUST be a valid tool call (tool\_call). Never respond with plain text.

6. Only one function call is allowed per turn. Never return multiple function calls in a single response. If multiple actions are needed, handle them sequentially across turns.
\\\\
\textbf{USER REQUEST:}
\\\\
\$USER\_SIMULATOR\_UTTERANCE
\end{prompt}
\caption{Prompt Template for the Modular-LLM Dialogue Manager, specifying task responsibilities, response rules, and interaction guidelines for subsystem coordination.}
    \label{fig:modllm_ds_prompt}
\end{figure*}
\begin{figure*}
  \centering
    \begin{prompt}
\\
\textbf{System Info}
\\\\
You are required to evaluate a task oriented dialogue on several metrics, including task completion, naturalness, coherence and dialogue-level diversity.
Alongside the dialogue, you are also provided with a user goal which states the specific requirement of the user.
\\\\
\textbf{TASK:}
\\\\
Here is some detailed explanations for the metrics:

1. Task completion
You should check whether each intention in the user goal is fulfilled in the conversation. The task is completed ONLY if all the intentions are fulfilled.
This would be a binary metric and you should only response with Yes or No.

This would be a binary metric and you should only response with Yes or No.
\\\\
2. Naturalness

This metric measures the resemblance to human.

In the dialogue, the user or the system could either be AI or human.

You should report a numeric rating from 1 to 5, where 5 represents most likely to be human.

You are required to evaluate the naturalness of both the user and the system.

Here are some more detailed guidelines for naturalness for your reference:

1: The speaker continuously repeat itself, typical robotic behavior. Or the speech is hard to understand.

2: The speaker repeat itself occasionally, the vocabulary is limited, like a robot.

3: The speaker does not have repeated behaviors (unless for verifying information). Vocabulary is enough to communicate effectively, speech is easy to understand. But I am confident that human rarely speak like this.

4: The speaker is likely to be a human. There is rarely logical inconsistency. But from some details I feel like the utterance is a bit weird and somewhat resembles AI.

5: Can not really tell if this is AI or human. Human could probably say the same thing in real life.
\\\\
3. Coherence

This metric measures the logical consistency within a dialogue.

You should report a numeric rating from 1 to 3, where 3 represents the best coherence.

Here is some detailed guidelines for coherence.

a. Locally, the utterances are coherent/logical based on previous turns of conversations.

b. Globally, the utterances reasonably and logically adhere to achieving the initial user goal step by step.

If both conditions a and b are satisfied, you should give a score of 3. If only one condition is satisfied, you should give a score of 2. Report 1 if none of the conditions are satisfied.
\\\\
4. Dialogue-level diversity

In addition to trying to achieve the initial goal, does the user introduce some reasonable deviations from the normal conversation flow?

Give a score from:

3 (highest score): > 20\% of the time (frequently deviate from normal flow of the conversation)

2: 0\% < deviation frequency < 20\% (Normal)

1 (lowest score): ~ 0\% (too artificial, maximizing information exchange)
\\\\
Note that for naturalness and coherence, you need to evaluate both the user and the system. For dialogue-level diversity, you only need to evaluate the user.

You should return 6 results in total, with the order of task completion, naturalness for the user, natualness for the system, coherence for the user, coherence for the system, diversity for the user.

Each evaluation results should be separated by commas. For example, 'Yes,5,3,3,1,2' will be a valid response.

Please be strict on the format of your response. Do not include any other words like 'Sure!', 'Here is the result:'. Simply response with only the results.
\\\\
The user goal is as following:

\$user\_goal
\\\\
The dialogue to be evaluated is as following:

\$dialogue

\end{prompt}

\caption{Prompt template for Dialogue Evaluation Task, describing detailed guidelines for assessing task completion, naturalness, coherence, and dialogue-level diversity.}
    \label{fig:dialoguequality}  
\end{figure*}

\begin{figure*}
  \centering
    \begin{prompt}
\\
\textbf{System Info}
\\\\
ROLE: You are an Intent Detection system designed to classify user requests into predefined domains and intents using the provided tool schema. You are NOT allowed to return free-form messages outside tool calls.
\\\\
\textbf{AVAILABLE INTENTS:}
\\\\
For all intent detections, use these exact names:

1. booking-request: User wants to proceed with the booking.

2. booking-success: The booking was successful and has some booking number.

3. booking-failure: There is a failure in the booking.

4. dbretrieval-request: User is looking for some information.

5. dbretrieval-success: The data is fetched from the DB and the retrieval was successful.

6. dbretrieval-failure: There is a failure in fetching the data from the DB.

7. detection-unknown: If the input doesn't fall into any of the above
\\\\
\textbf{AVAILABLE DOMAINS:}
\\\\
1. Classify the request into only one of the following domains (choose the closest match):
   * restaurant, hotel, train

2. Not all utterances can be categorized into a domain. In such cases, use "donotcare".
\\\\
\textbf{TASK:}
\\\\
1. Analyze the provided input. Dialogue history is provided to understand the context better.

2. Classify the request into only one of the above predefined intents (the closest match) and domain

3. Return the detected intent and domain by using those exact names.

4. Do not add any other information or explanation or comments.

5. Every response MUST be a valid tool call (tool\_call). Never respond with plain text.

6. Only one function call is allowed per turn.
\\\\
\textbf{INPUT:}
\\\\
\$USER\_SIMULATOR\_UTTERANCE
\end{prompt}
\caption{Prompt template for the Intent Detection module, specifying task guidelines for classifying user requests into predefined intents and domains.}
    \label{fig:modprog_intent_detection}
\end{figure*}

\begin{figure*}
  \centering
    \begin{prompt}
\\
\textbf{System Info}
\\\\
ROLE: You are an Slot Extraction system designed to identify and extract key entities from user requests to support downstream tasks using the provided tool schema. You are NOT allowed to return free-form messages outside tool calls.
\\\\
\textbf{TASK:}
\\\\
1. Analyze the provided user request.

2. Identify and extract relevant slots (e.g., name, area, time, date, type of cuisine, number of people, type of hotel) based on the task context.

3. Focus on extracting the most concise and precise values for each slot, avoiding unnecessary descriptive phrases or additional words.

4. Return the extracted slots in a structured format.

5. Return only the formatted data—do not add explanations, comments, or additional information.

6. Only extract a slot if it is **explicitly mentioned** in the user input. Do not infer, assume, or hallucinate values based on common patterns or prior examples.

7. If a relevant slot is not present in the input, **omit it from the output entirely**—do not fabricate or guess.

8. When handling follow-up user requests, compare the new input to the dialogue history:  

   - If a slot was **previously extracted** but the new input **replaces or contradicts** it (e.g., rephrasing or simplifying),  

     then **explicitly reset** that slot by setting its value to an empty string (e.g., `"area": ""`).  

   - If the user is merely **adding new information** (e.g., number of people, dates), do **not** reset previous values—**preserve them**.
\\\\
\textbf{RESPONSE RULES:}
\\\\
1. Use the most appropriate function call based on the user’s request and available data.

2. Every response MUST be a valid tool call (tool\_call). Never respond with plain text.

3. Only one function call is allowed per turn.
\\\\
\textbf{USER REQUEST:}
\\\\
\$USER\_SIMULATOR\_UTTERANCE
\end{prompt}
\caption{Prompt template for the Slot Extraction module, outlining task instructions for identifying and extracting structured key entities from user requests.}
    \label{fig:modprog_slot_extraction}
\end{figure*}

\begin{figure*}
  \centering
    \begin{prompt}
\\
\textbf{System Info}
\\\\
ROLE: You are a Response Generation system responsible for crafting contextually appropriate and concise replies based strictly on the provided input using the provided tool schema. You are NOT allowed to return free-form messages outside tool calls.
\\\\
\textbf{TASK:}
\\\\
Given the input data (domain, intent, extracted slots, database (DB) information, and dialogue history):

1. Generate a meaningful response:

   a. If additional information is required to proceed, respond conversationally using direct and focused phrasing.

   b. If recommendations are provided in the DB:

      * Ask the user to choose from the list of options.

      * Clearly present all options to the user for selection. Do not decide on any recommendation yourself.

   c. For train bookings, if the database does not have trains available at the exact requested time, clarify with the user whether they are interested in seeing the closest available options that best match their query.
\\\\
2. Guidelines for Response:
\\\\
   a. Responses must be concise and to the point.
   
   b. Avoid unnecessary elaboration or an overly conversational tone.
   
   c. Do not generate or fabricate any information that is not explicitly present in the DB or provided input.
   
   d. If too many records are available, the database system returns only the first five. If the required information is not available in the returned records, request the user for additional information to narrow down the results.
   
   e. Do not generate any booking confirmation (reference number) on your own.
   
   f. If the input contains booking confirmation (reference number), share the same to user without any changes.
   
   g. Every response MUST be a valid tool call (tool\_call). Never respond with plain text.
   
   h. Only one function call is allowed per turn. Never return multiple function calls in a single response. If multiple actions are needed, handle them sequentially across turns.
\\\\
\textbf{INPUT:}
\\\\
\$USER\_SIMULATOR\_UTTERANCE
\end{prompt}
\caption{Prompt template for the Response Generation module, specifying task and response guidelines for producing contextually appropriate and structured replies based on user input and dialogue history.}
    \label{fig:modprog_response_generation}
\end{figure*}
\clearpage

\begin{nolinenumbers}
\begin{minipage}{\textwidth}
\begin{lstlisting}[language=json]
[
        {
            "type": "function",
            "function": {
                "name": "followup",
                "description": "Use this function to respond to the user with follow-up messages. This includes  asking for missing or unclear information, confirming details, sharing booking reference numbers, or continuing the dialogue based on the current conversation state.",
                "parameters": {
                    "type": "object",
                    "properties": {
                        "message": {
                            "type": "string",
                            "description": "The response from the dialogue system to the user"
                        }
                    },
                    "required": ["message"],
                    "additionalProperties": false
                }
            }
        },
        {
            "type": "function",
            "function": {
                "name": "retrievefromrestaurantdb",
                "description": "Use this function to query the restaurant database and retrieve restaurants that match optional filters such as area, pricerange, food (cuisine), or restaurant name. This function is typically used to find available restaurant options before validating or making a reservation. Returns up to 5 matching restaurants, or fewer if less than 5 matches are found.",
                "parameters": {
                    "type": "object",
                    "properties": {
                        "area": {
                            "type": "string",
                            "enum": ["centre", "north", "east", "west", "south"],
                            "description": "The area/location/place of the restaurant. Optional."
                        },
                        "pricerange": {
                            "type": "string",
                            "enum": ["cheap", "moderate", "expensive"],
                            "description": "The price budget for the restaurant. Optional."
                        },
                        "food": {
                            "type": "string",
                            "description": "The cuisine of the restaurant you are looking for. Optional."
                        },
                        "name": {
                            "type": "string",
                            "description": "The name of the restaurant. Optional."
                        }
                    },
                    "required": []
                }
            }
        }                                          
]
\end{lstlisting}
\captionof{figure}{Schema definition detailing parameters and constraints for querying the database and booking confirmation (Part 1/6).}
\label{lst:tool-schema-1}
\end{minipage}

\clearpage
\begin{minipage}{0.95\textwidth}
\begin{lstlisting}[language=json]
[
        {
            "type": "function",
            "function": {
                "name": "retrievefromhoteldb",
                "description": "Use this function to query the hotel database and retrieve hotels/guesthouses that match optional filters such as area, pricerange, type, hotel name, internet, parking, or stars. This function is typically used to find available hotel options before validating or making a reservation. Returns up to 5 matching hotels, or fewer if less than 5 matches are found.",
                "parameters": {
                    "type": "object",
                    "properties": {
                        "area": {
                            "type": "string",
                            "enum": ["centre", "north", "east", "west", "south"],
                            "description": "The area/location/place of the hotel. Optional."
                        },
                        "pricerange": {
                            "type": "string",
                            "enum": ["cheap", "moderate", "expensive"],
                            "description": "The price budget for the hotel. Optional."
                        },
                        "type": {
                            "type": "string",
                            "enum": ["hotel", "guesthouse"],
                            "description": "What is the type of the hotel. Optional."
                        },
                        "name": {
                            "type": "string",
                            "description": "The name of the hotel. Optional."
                        },
                        "internet": {
                            "type": "string",
                            "enum": ["yes", "no"],
                            "description": "Indicates, whether the hotel has internet/wifi or not. Optional."
                        },
                        "parking": {
                            "type": "string",
                            "enum": ["yes", "no"],
                            "description": "Indicates, whether the hotel has parking or not. Optional."
                        },
                        "stars": {
                            "type": "object",
                            "description": "The star rating of the hotel. Optional.",
                            "properties": {
                                "operator": { "type": "string", "enum": ["=", ">=", "<=", ">", "<"] },
                                "value": { "type": "string", "enum": ["1", "2", "3", "4", "5"] }
                            },
                            "required": ["operator", "value"],
                            "additionalProperties": false
                        }
                    },
                    "required": []
                }
            }
        }                                     
]
\end{lstlisting}
\captionof{figure}{Schema definition detailing parameters and constraints for querying the database and booking confirmation (Part 2/6).}
\label{lst:tool-schema-2}
\end{minipage}

\clearpage
\begin{minipage}{0.95\textwidth}
\begin{lstlisting}[language=json]
[
        {
            "type": "function",
            "function": {
                "name": "retrievefromtraindb",
                "description": "Use this function to query the train database and retrieve trains that match optional filters such as destination, departure, day, arriveby, or leaveat. This function is typically used to find available options before validating or making a reservation. Returns up to 5 matching trains, or fewer if less than 5 matches are found.",
                "parameters": {
                    "type": "object",
                    "properties": {
                        "destination": {
                            "type": "string",
                            "description": "Destination of the train. Optional."
                        },
                        "departure": {
                            "type": "string",
                            "description": "Departure location of the train. Optional."
                        },
                        "day": {
                            "type": "string",
                            "enum": ["monday", "tuesday", "wednesday", "thursday", "friday", "saturday", "sunday"],
                            "description": "Journey day of the train. Optional."
                        },
                        "arriveby": {
                            "type": "object",
                            "description": "Arrival time of the train. Optional.",
                            "properties": {
                                "operator": { "type": "string", "enum": ["=", ">=", "<=", ">", "<"] },
                                "value": { "type": "string", "pattern": "^(0[0-9]|1[0-9]|2[0-3]):[0-5][0-9]$",
                                           "description": "A time string formatted as HH:MM (24-hour format)."
                                         }
                            },
                            "required": ["operator", "value"],
                            "additionalProperties": false
                        },
                        "leaveat": {
                            "type": "object",
                            "description": "Leaving time for the train. Optional.",
                            "properties": {
                                "operator": { "type": "string", "enum": ["=", ">=", "<=", ">", "<"] },
                                "value": { "type": "string", "pattern": "^(0[0-9]|1[0-9]|2[0-3]):[0-5][0-9]$",
                                           "description": "A time string formatted as HH:MM (24-hour format)."
                                         }
                            },
                            "required": ["operator", "value"],
                            "additionalProperties": false
                        }
                    },
                    "required": []
                }
            }
        }                                   
]
\end{lstlisting}
\captionof{figure}{Schema definition detailing parameters and constraints for querying the database and booking confirmation (Part 3/6).}
\label{lst:tool-schema-3}
\end{minipage}

\clearpage
\begin{minipage}{0.95\textwidth}
\begin{lstlisting}[language=json]
[
    {
        "type": "function",
        "function": {
            "name": "validaterestaurantbooking",
            "description": "Use this function to check the availability of a restaurant based on user preferences such as area, food (cuisine), pricerange, name, people, day, and time before proceeding with a reservation. This function should be called to validate whether a booking can be made with the provided details. If the details are accurate, it returns a booking reference number.",
            "parameters": {
                "type": "object",
                "properties": {
                    "area": {
                        "type": "string",
                        "enum": ["centre", "north", "east", "west", "south"],
                        "description": "The area/location/place of the restaurant."
                    },
                    "pricerange": {
                        "type": "string",
                        "enum": ["cheap", "moderate", "expensive"],
                        "description": "The price budget for the restaurant."
                    },
                    "food": {
                        "type": "string",
                        "description": "The cuisine of the restaurant you are looking for."
                    },
                    "name": {
                        "type": "string",
                        "description": "The name of the restaurant."
                    },
                    "phone": {
                        "type": "string",
                        "description": "Phone number of the restaurant. Optional."
                    },
                    "postcode": {
                        "type": "string",
                        "description": "Postal code of the restaurant. Optional."
                    },
                    "address": {
                        "type": "string",
                        "description": "Address of the restaurant. Optional."
                    },
                    "people": {
                        "type": "string",
                        "enum": ["1", "2", "3", "4", "5", "6", "7", "8"],
                        "description": "Number of people for the restaurant reservation."
                    },
                    "day": {
                        "type": "string",
                        "enum": ["monday", "tuesday", "wednesday", "thursday", "friday", "saturday", "sunday"],
                        "description": "Day of the restaurant reservation."
                    },
                    "time": {
                        "type": "string",
                        "pattern": "^(0[0-9]|1[0-9]|2[0-3]):[0-5][0-9]$",
                        "description": "Time of the restaurant reservation, formatted as HH:MM (24-hour format)."
                    },                    
                },
                "required": ["food", "area", "pricerange", "name", "people", "day", "time"],
                "additionalProperties": false
            }
        }
    }                               
]
\end{lstlisting}
\captionof{figure}{Schema definition detailing parameters and constraints for querying the database and booking confirmation (Part 4/6).}
\label{lst:tool-schema-4}
\end{minipage}

\clearpage
\begin{minipage}{0.95\textwidth}
\begin{lstlisting}[language=json]
[
    {
        "type": "function",
        "function": {
            "name": "validatehotelbooking",
            "description": "Use this function to check the availability of a hotel based on user preferences such as area, type (hotel/guesthouse), pricerange, name, internet, parking, stars, people, day and stay before proceeding with a reservation. This function should be called to validate whether a booking can be made with the provided details. If the details are accurate, it returns a booking reference number.",
            "parameters": {
                "type": "object",
                "properties": {
                    "area": {
                        "type": "string",
                        "enum": ["centre", "north", "east", "west", "south"],
                        "description": "The area/location/place of the hotel."
                    },
                    "pricerange": {
                        "type": "string",
                        "enum": ["cheap", "moderate", "expensive"],
                        "description": "The price budget for the hotel."
                    },
                    "type": {
                        "type": "string",
                        "enum": ["hotel", "guesthouse"],
                        "description": "What is the type of the hotel."
                    },
                    "name": {
                        "type": "string",
                        "description": "The name of the hotel."
                    },
                    "internet": {
                        "type": "string",
                        "enum": ["yes", "no"],
                        "description": "Indicates, whether the hotel has internet/wifi or not."
                    },
                    "parking": {
                        "type": "string",
                        "enum": ["yes", "no"],
                        "description": "Indicates, whether the hotel has parking or not."
                    },
                    "stars": {
                        "type": "string",
                        "enum": ["1", "2", "3", "4", "5"],
                        "description": "The star rating of the hotel."
                    },
                    "people": {
                        "type": "string",
                        "enum": ["1", "2", "3", "4", "5", "6", "7", "8"],
                        "description": "Number of people for the hotel booking."
                    },
                    "day": {
                        "type": "string",
                        "enum": ["monday", "tuesday", "wednesday", "thursday", "friday", "saturday", "sunday"],
                        "description": "Day of the hotel booking."
                    },                    
                    "stay": {
                        "type": "string",
                        "enum": ["1", "2", "3", "4", "5", "6", "7", "8"],
                        "description": "Length of stay at the hotel."
                    },                    
                    "phone": {
                        "type": "string",
                        "description": "Phone number of the hotel. Optional."
                    },
                    "postcode": {
                        "type": "string",
                        "description": "Postal code of the hotel. Optional."
                    },
                    "address": {
                        "type": "string",
                        "description": "Address of the hotel. Optional."
                    }                    
                },
                "required": ["area", "pricerange", "type", "internet",
                              "parking", "name", "stars", "people", "day", "stay"],
                "additionalProperties": false
            }
        }
    }                              
]
\end{lstlisting}
\captionof{figure}{Schema definition detailing parameters and constraints for querying the database and booking confirmation (Part 5/6).}
\label{lst:tool-schema-5}
\end{minipage}

\clearpage
\begin{minipage}{0.95\textwidth}
\begin{lstlisting}[language=json]
[
        {
            "type": "function",
            "function": {
                "name": "validatetrainbooking",
                "description": "Use this function to check the availability of a train based on user preferences such as destination, departure, arriveby, leaveat, day, people, and trainid before proceeding with a reservation. This function should be called to validate whether a booking can be made with the provided details. If the details are accurate, it returns a booking reference number.",
                "parameters": {
                    "type": "object",
                    "properties": {
                        "destination": {
                            "type": "string",
                            "description": "Destination of the train."
                        },
                        "departure": {
                            "type": "string",
                            "description": "Departure location of the train."
                        },
                        "day": {
                            "type": "string",
                            "enum": ["monday", "tuesday", "wednesday", "thursday", "friday", "saturday", "sunday"],
                            "description": "Journey day of the train."
                        },
                        "arriveby": {
                            "type": "string",
                            "pattern": "^(0[0-9]|1[0-9]|2[0-3]):[0-5][0-9]$",
                            "description": "Arrival time of the train."
                        },
                        "leaveat": {
                            "type": "string",
                            "pattern": "^(0[0-9]|1[0-9]|2[0-3]):[0-5][0-9]$",
                            "description": "Leaving time for the train."
                        },
                        "people": {
                            "type": "string",
                            "enum": ["1", "2", "3", "4", "5", "6", "7", "8"],
                            "description": "Number of train tickets for the booking."
                        },
                        "trainid": {
                            "type": "string",
                            "description": "ID of the train."
                        },
                        "price": {
                            "type": "string",
                            "description": "Price of the train journey. Optional."
                        },
                        "duration": {
                            "type": "string",
                            "description": "Duration of the travel. Optional."
                        }
                    },
                    "required": ["destination", "departure", "day", "arriveby", "leaveat", "people", "trainid"],
                    "additionalProperties": false
                }
            }
        }                              
]
\end{lstlisting}
\captionof{figure}{Schema definition detailing parameters and constraints for querying the database and booking confirmation (Part 6/6).}
\label{lst:tool-schema-6}
\end{minipage}
\end{nolinenumbers}
\end{document}